\documentclass{article}

\PassOptionsToPackage{square, numbers, comma, sort&compress}{natbib}
\usepackage[preprint]{neurips_2022_modified}  % to compile a preprint version

\usepackage{xr-hyper}
\usepackage{amsfonts}
\usepackage{amsmath}
\usepackage{amssymb}  % other imports include: \usepackage{mathtools} and \usepackage{MnSymbol}
\usepackage{booktabs}
\usepackage{graphicx}
\usepackage[font=footnotesize, margin=0pt, skip=4pt, singlelinecheck=off]{caption}
\usepackage[font=footnotesize, justification=centering, margin=0pt, skip=0pt]{subcaption}
\usepackage{dirtytalk}
\usepackage{enumitem}\setlist{leftmargin=15mm}
\usepackage[T1]{fontenc}
\usepackage{hyperref}
\usepackage{url}
\usepackage[utf8]{inputenc}
\usepackage{microtype}
\usepackage{nicefrac}
\usepackage{soul}
\usepackage[dvipsnames]{xcolor}
\usepackage[breakable, theorems, skins]{tcolorbox}

\newcommand{\bx}{\mathbf{x}}
\newcommand{\bz}{\mathbf{z}}

\newcommand{\bbE}{\mathbb{E}}

\newcommand{\cL}{\mathcal{L}}
\newcommand{\cN}{\mathcal{N}}

\newcommand{\cX}{\mathcal{X}}
\newcommand{\KL}{K\!L}

\newcommand{\cZ}{\mathcal{Z}}
\newcommand{\R}{\mathbb{R}}
\newcommand\DDS{\displaystyle}
\newcommand\hQ{\widehat{Q}}
\newcommand\hA{\widehat{A}}
\newcommand\tpB{\tilde{p}_B}
\DeclareMathOperator*{\E}{\mathbb{E}}
\DeclareMathOperator*{\var}{var}
\definecolor{PltBlue}{HTML}{1f77b4}
\definecolor{PltOrange}{HTML}{ff7f0e}
\definecolor{PltGreen}{HTML}{2ca02c}
\definecolor{PltRed}{HTML}{d62728}
\definecolor{PltPurple}{HTML}{9467bd}
\definecolor{PltBrown}{HTML}{8c564b}
\definecolor{PltPink}{HTML}{e377c2}
\definecolor{PltGray}{HTML}{7f7f7f}
\definecolor{PltLime}{HTML}{bcbd22}
\definecolor{PltCyan}{HTML}{17becf}
\newtheorem{prop}{Proposition}[section]
\newtheorem{coro}{Corollary}[section]

\title{Designing losses for data-free training\\ of normalizing flows on Boltzmann distributions}
\author{%
    Loris Felardos$^{1,2}$ \\
    \texttt{\small loris.felardos.212@use.startmail.com} \\
    \\
    \And
    Jérôme H\'enin$^{2}$ \\
    \texttt{\small jerome.henin@cnrs.fr} \\
    \\
    \And
    Guillaume Charpiat$^{1}$ \\
   \texttt{\small guillaume.charpiat@inria.fr} \\
}

\makeatletter
\newcommand*{\addFileDependency}[1]{% argument=file name and extension
  \typeout{(#1)}
  \@addtofilelist{#1}
  \IfFileExists{#1}{}{\typeout{No file #1.}}
}
\makeatother
\begin{document}
\maketitle

{\footnotesize\centering{    ${}^{1}$ Universit\'e Paris-Saclay, CNRS, Inria, Laboratoire interdisciplinaire des sciences du num\'erique, Orsay, France\\
    ${}^2$ Universit\'e Paris Cit\'e, CNRS, Laboratoire de Biochimie Th\'eorique UPR 9080, Paris, France\\
}}
\vspace{10mm}

%%%%%%%%%%%%%%%%%%%%%%%%%%%%%%%%%%%%%%%%%%%%%%%%%%%%%%%%%%%%%%%%%%%%%%%%%%%%%%%%%%%%%%%%%%%%%%%%%%%
%%%%%%%%%%%%%%%%%%%%%%%%%%%%%%%%%%%%%%%%%%%%%%%%%%%%%%%%%%%%%%%%%%%%%%%%%%%%%%%%%%%%%%%%%%%%%%%%%%%
\begin{abstract}
    Generating a Boltzmann distribution in high dimension has recently been achieved with Normalizing Flows, which enable fast and exact computation of the generated density, and thus unbiased estimation of expectations. However, current implementations rely on accurate training data, which typically comes from computationally expensive simulations.
    There is therefore a clear incentive to train models with incomplete or no data by relying solely on the target density, which can be obtained from a physical energy model (up to a constant factor). For that purpose, we analyze the properties of standard losses based on Kullback-Leibler divergences. We showcase their limitations, in particular a strong propensity for mode collapse during optimization on high-dimensional distributions. We then propose strategies to alleviate these issues, most importantly a new loss function well-grounded in theory and with suitable optimization properties. Using as a benchmark the generation of 3D molecular configurations, we show on several tasks that, for the first time, imperfect pre-trained models can be further optimized in the absence of training data.
\end{abstract}

%%%%%%%%%%%%%%%%%%%%%%%%%%%%%%%%%%%%%%%%%%%%%%%%%%%%%%%%%%%%%%%%%%%%%%%%%%%%%%%%%%%%%%%%%%%%%%%%%%%
%%%%%%%%%%%%%%%%%%%%%%%%%%%%%%%%%%%%%%%%%%%%%%%%%%%%%%%%%%%%%%%%%%%%%%%%%%%%%%%%%%%%%%%%%%%%%%%%%%%
\section{Introduction}
\label{sec:intro}

\paragraph{Application context.}
In statistical physics, the properties of materials and molecular systems are expressed as expectations over probability distributions of microscopic configurations, which are determined by macroscopic, thermodynamic parameters.
Such expectations can be estimated numerically by Monte Carlo averaging using samples from physically relevant distributions, particularly the Boltzmann distribution characterizing systems at equilibrium with a thermostat.
The Boltzmann distribution over configurations $x$ is characterized by the density $p_B$, which is related to the potential energy $U_B$ by:
\begin{equation}
    p_B(x) = \frac{1}{\cZ_B} \cdot e^{-\beta U_B(x)}
\end{equation}
where $\beta = 1/k_B T$ is the inverse temperature, and $\cZ_B$ is a normalization factor known as the partition function.
Though there are usually closed form expressions or robust numerical methods to estimate $\tpB := \cZ_B\, p_B$, there is no direct method to sample it.
In practice, sampling is commonly performed with stochastic simulations of physical systems, however $p_B$ is typically high-dimensional and multimodal, so that simulations are plagued by long autocorrelation times.
Sampling with generative models, which produce i.i.d.~samples, is a potential avenue to overcome these limitations.

\paragraph{Normalizing Flows for Boltzmann distributions.}
Flow-based models (often just called normalizing flows) are a valuable type of architecture for this purpose (\cite{tabak2010densityestimation},~\cite{Tabak2013NonparametricDensityEstimation},~\cite{Rezende2015VariationalInferencewithNFs} and~\cite{Papamakarios2019NFsReview} for an overview), which is invertible and yields not only samples $x$ but also the probability density $p_G(x)$ of the generated distribution.
This in turns allows for unbiased estimation of expectations with respect to the ground-truth Boltzmann distribution via reweighting:
\begin{equation}
\underset{x \sim p_B}{\bbE}  [ f(x) ] = \underset{x \sim p_G}{\bbE} \left [ \frac{p_B(x)}{p_G(x)} \; f(x) \right ]
\end{equation}
for any function $f$, assuming that $p_G$ is nonzero over the support of $f$.
This is the case of Boltzmann generators~\cite{Noe2019}, which are based on Normalizing Flows with affine coupling layers~\cite{Dinh2016RealNVP}, trained to generate a known Boltzmann distribution.

Designing more robust and expressive normalizing flow architectures is an active field of research, with innovations such as rank-one perturbations to train fully connected layers~\cite{Kramer2010PPPP}, Augmented Normalizing Flows~\cite{Huang2020AugmentedNFs}, Stochastic Normalizing Flows~\cite{Wu2020StochasticNFs}, Smooth Normalizing Flows~\cite{Kohler2021SmoothNFs} and base distribution resampling~\cite{Stimper2021ResamplingBaseDistributions}.

\paragraph{Towards data-free training.}
In principle, a loss function based on a well-chosen $\KL$ divergence should allow for the training of normalizing flows in the absence of data, merely based on the knowledge of the target Boltzmann distribution up to a constant factor~\cite{Papamakarios2019NFsReview}.
However, there are no claims of successful numerical experiments in the literature, suggesting that this approach may be impractical for so-far undocumented reasons.
Thus, it remains that in practice, Boltzmann generators must be trained using accurate reference data, which makes them applicable to systems that have already been sampled by other means, rather than standalone substitutes to simulations for studying new, unknown systems.
Generally speaking, training generative models on high-dimensional distributions is difficult because it puts a high demand on the space to be covered during training; training them in the absence of complete reference data is to date an unsolved problem.
There are two requirements for success: proper convergence (which implies stability of the generated distribution near its target), and exploration of the ground-truth distribution.
Here we focus on stability and propose the very first data-free loss leading to stable training.
We discuss possible approaches for an exploration strategy in the discussion (section~\ref{sec:conclusion}).

\paragraph{Contributions and overview.}
In this work, we analyze the properties of loss functions based on Kullback-Leibler divergences, and showcase their limitations, in particular their lack of robustness with respect to discretization, with a general tendency towards mode drop that makes data-free training unstable.
We then introduce a loss function that exhibits stable refinement training in the absence of data, after an initial data-dependent pre-training.
We assess all losses and training strategies on a toy model (a high-dimensional double-well potential) and two molecular systems.
We further discuss the sensitivity of normalizing flow training to degrees of freedom with broad probability distributions in the output, and propose strategies to avoid these effects at the level of the training criterion, without added architectural constraints such as equivariance or invariance.

%%%%%%%%%%%%%%%%%%%%%%%%%%%%%%%%%%%%%%%%%%%%%%%%%%%%%%%%%%%%%%%%%%%%%%%%%%%%%%%%%%%%%%%%%%%%%%%%%%%
%%%%%%%%%%%%%%%%%%%%%%%%%%%%%%%%%%%%%%%%%%%%%%%%%%%%%%%%%%%%%%%%%%%%%%%%%%%%%%%%%%%%%%%%%%%%%%%%%%%
\section{Optimizing Kullback-Leibler Divergences}
\label{sec:KLs}

\paragraph{$\KL$ divergence defined in z-space.}
The goal of training a Normalizing Flow $G = F^{-1}$ is to obtain a one-to-one mapping between a known base distribution $q_\cN$ (typically Gaussian) and a target distribution $p_B$, such that the pushforward measure $p_G$ of $q_\cN$ by $G$ is similar to $p_B$.
\begin{align*}
    \overbrace{z_\cN \sim q_\cN}^\text{Gaussian distribution} & \boldsymbol{\overset{G}{\xrightarrow{\hspace{2.5cm}}}} \overbrace{x_G = G(z_\cN) \sim p_G}^\text{generated distribution} \\
    z_F = F(x_B) \sim q_F & \boldsymbol{\underset{F}{\xleftarrow{\hspace{2.5cm}}}} \underbrace{ x_B \sim p_B}_\text{target distribution}
\end{align*}

Since normalizing flows are bijective, the conventional way of achieving this is by providing $x_B$ samples from $p_B$ (usually from a dataset) to the inverse function $F$ and then minimizing the $\KL$ divergence between the pushforward measure $q_F$ of $p_B$ by $F$ and the known $q_\cN$.
\begin{subequations}
    \label{eq:KLs:KLz_dvt}
    \begin{align}
        \KL(q_F||& q_\cN) = \int q_F(z) \log \frac{q_F(z)}{q_\cN(z)} \ dz \label{subeq:KLs:KLz_dvt_a} \\
        & = \log\cZ_\cN - S_B + \underset{x_B \sim p_B}{\bbE} \left[ \frac{1}{2\sigma^2} U_\cN(F(x_B)) - \log \left| \det \left( \frac{\partial F(x_B)}{\partial x_B} \right) \right| \right] \label{subeq:KLs:KLz_dvt_b}
    \end{align}
\end{subequations}

When leveraging the principle of Stochastic Gradient Descent, this gives rise to the following standard and data-\textit{dependent} loss function, with $\bx_B$ a mini-batch of $x_B$ points sampled from $p_B$ (appendix~\ref{sec:appendix:KLz}):
\begin{equation}
    \label{eq:KLs:KLz_loss}
    \cL_{\KL z}(\bx_B) = \sum_{i=1}^n \left[ \frac{1}{n} \cdot \left[ \frac{1}{2\sigma^2} U_\cN(F(x_{B,i})) - \log \left| \det \left( \frac{\partial F(x_{B,i})}{\partial x_{B,i}} \right) \right| \right] \right]
\end{equation}

\paragraph{$\KL$ divergence defined in x-space.}
Another loss can be derived in an almost identical fashion by defining a $\KL$ divergence in x-space instead (appendix~\ref{sec:appendix:KLx}):
\begin{subequations}
    \label{eq:KLs:KLx_dvt}
    \begin{align}
        \KL(p_G||p_B) & = \int p_G(x) \log \frac{p_G(x)}{p_B(x)} \ dx \label{subeq:KLs:KLx_dvt_a} \\
        & = \log\cZ_B - S_\cN + \underset{z_\cN \sim q_\cN}{\bbE} \left[ \beta U_B(G(z_\cN)) - \log \left| \det \left( \frac{\partial G(z_\cN)}{\partial z_\cN} \right) \right| \right] \label{subeq:KLs:KLx_dvt_b}
    \end{align}
\end{subequations}

This leads to the following data-\textit{free} loss function (with $\bz_\cN$ a mini-batch of $z_\cN$ points)::
\begin{equation}
    \label{eq:KLs:KLx_loss}
    \cL_{\KL x}(\bz_\cN) = \sum_{i=1}^n \left[ \frac{1}{n} \cdot \left[ \beta U_B(G(z_{\cN,i})) - \log \left| \det \left( \frac{\partial G(z_{\cN,i})}{\partial z_{\cN,i}} \right) \right| \right] \right]
\end{equation}

\paragraph{Comparing $\cL_{\KL x}$ with $\cL_{\KL z}$.}
When optimizing over mini-batches, these two loss functions behave very differently. $\cL_{\KL z}$ is known to be very stable and leads to good performance~\cite{Papamakarios2019NFsReview} while $\cL_{\KL x}$ is more erratic and often leads to mode collapse. To illustrate this, we pre-train a model on a simple dataset with $\cL_{\KL z}$ and then fine-tune it with $\cL_{\KL x}$. This is the general experimental setup used in this work. Poor pre-trainings are allowed as long as they do not miss an entire mode of the target distribution so as to analyze whether the fine-tunings manage to refine $p_G$ successfully.
See section~\ref{sec:conclusion} on possible strategies to remove this data-dependent pre-training in the future.

\begin{figure}[!b]
    \centering
    \begin{subfigure}{.30\textwidth}\hspace{5pt}
        \includegraphics[width=102pt, height=74pt]{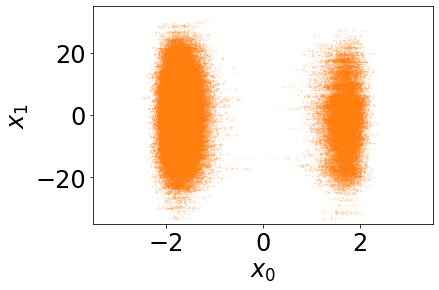}
        \caption{2D Projection of the dataset: \texttt{Double Well 12D}}
        \label{fig:PP_double_well_12Dthick_data_scatter}
    \end{subfigure}
    \begin{subfigure}{.68\textwidth}\hspace{12pt}
        \includegraphics[width=240pt, height=68pt]{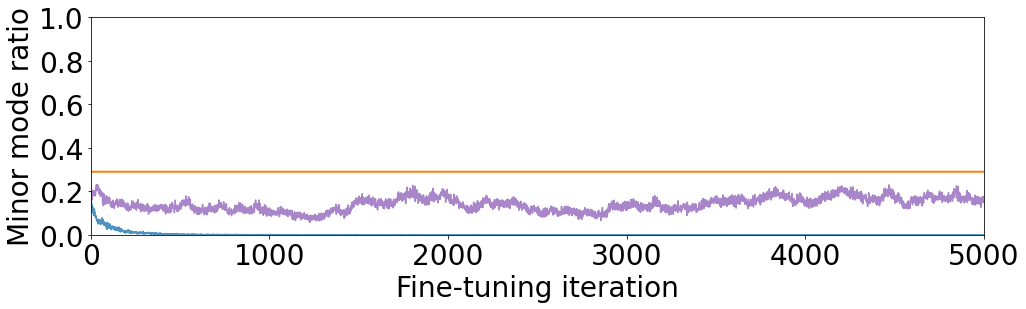}
        \vspace{5pt}
        \caption{Percentage of generated samples $x_G \sim p_G$ in the minor mode during fine-tunings compared to the real ratio from $p_B$ (in \textcolor{PltOrange}{\ul{orange}}).}
        \label{fig:PP_double_well_12Dthick_KLx_ratios}
    \end{subfigure}
    \begin{subfigure}{.24\textwidth}\centering
        \includegraphics[width=94pt, height=74pt]{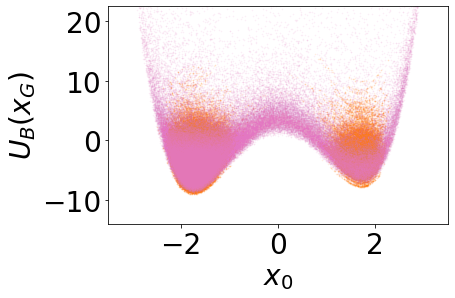}
        \caption{Partial pre-training.}
        \label{fig:PP_double_well_12Dthick_pretraining_v1_x0UB}
    \end{subfigure}
    \begin{subfigure}{.24\textwidth}\centering
        \includegraphics[width=94pt, height=74pt]{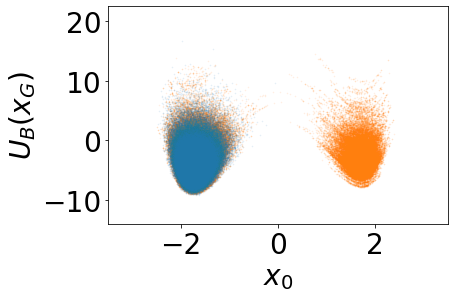}
        \caption{Fine-tuning of~\ref{fig:PP_double_well_12Dthick_pretraining_v1_x0UB}.}
        \label{fig:PP_double_well_12Dthick_KLx_v1_x0UB}
    \end{subfigure}
    \begin{subfigure}{.24\textwidth}\centering
        \includegraphics[width=94pt, height=74pt]{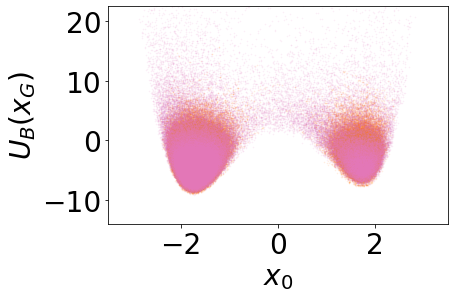}
        \caption{Complete pre-training.}
        \label{fig:PP_double_well_12Dthick_pretraining_v2_x0UB}
    \end{subfigure}
    \begin{subfigure}{.24\textwidth}\centering
        \includegraphics[width=94pt, height=74pt]{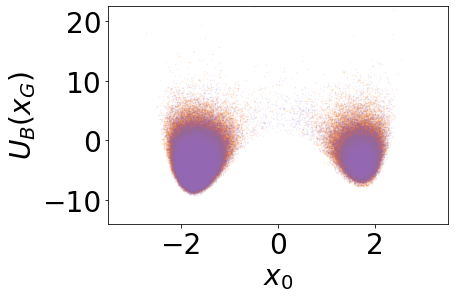}
        \caption{Fine-tuning of~\ref{fig:PP_double_well_12Dthick_pretraining_v2_x0UB}.}
        \label{fig:PP_double_well_12Dthick_KLx_v2_x0UB}
    \end{subfigure}
    \caption{
        Results of two fine-tunings with $\cL_{\KL x}$ after pre-trainings of different lengths with $\cL_{\KL z}$. Data from $p_B$ (i.e. the dataset) is represented in \textcolor{PltOrange}{\ul{orange}}. Both pre-training results are represented in \textcolor{PltPink}{\ul{pink}}. Fine-tuning results after the partial pre-training are represented in \textcolor{PltBlue}{\ul{blue}} (note the total collapse to the major mode). Fine-tuning results after the complete pre-training are represented in \textcolor{PltPurple}{\ul{purple}}. Figures~\ref{fig:PP_double_well_12Dthick_pretraining_v1_x0UB} to \ref{fig:PP_double_well_12Dthick_KLx_v2_x0UB} all represent the potential energy $U_B$ of generated samples $x_G$ (in ordinates) as a function of the multi-modal dimension (in abscissa).
    }
    \label{fig:PP_double_well_12Dthick}
\end{figure}

The dataset is a simple double well in 12 dimensions similar to those used in previous works\cite{Noe2019, Midgley2021} (figure~\ref{fig:PP_double_well_12Dthick_data_scatter}), where the first dimension is bimodal and the 11 other dimensions are independent and Gaussian with a standard deviation of 10. The lack of normalization is intended to exhibit how difficult this task already is for $\cL_{\KL x}$. Even with a partial pre-training that already samples the bottom of each mode (figure~\ref{fig:PP_double_well_12Dthick_pretraining_v1_x0UB}), it is incapable of keeping both modes and collapses to the main one (figure~\ref{fig:PP_double_well_12Dthick_KLx_v1_x0UB}). A complete pre-training with $\cL_{\KL z}$ (figure~\ref{fig:PP_double_well_12Dthick_pretraining_v2_x0UB}) results in a better fine-tuning (figure~\ref{fig:PP_double_well_12Dthick_KLx_v2_x0UB}) but is still not sufficient to completely stabilize $\cL_{\KL x}$ as shown in figure~\ref{fig:PP_double_well_12Dthick_KLx_ratios} when looking at the ratio between the modes over time. Note that this failure is \textit{not} due to the poor normalization of the target distribution, which only exacerbates this undesirable behavior, since $\cL_{\KL x}$ also fails on more complex datasets with good normalization (appendix~\ref{sec:appendix:KLx_dialanine_collapse}).

\paragraph{Making $\cL_{\KL z}$ data-free.}
The standard loss $\cL_{\KL z}$ cannot be used in a data-free setting since it relies on samples from $p_B$, but it can be modified to use samples from $p_G$ instead by leveraging importance sampling (appendix~\ref{sec:appendix:KLz_bws}):
\begin{equation}
    \label{eq:KLs:KLzbw_loss}
    \cL_{\KL z}^\mathrm{df}(\bx_G^\ddagger) = \sum_{i=1}^n \frac{1}{n} \cdot \left[ \left( \frac{\tpB(x_G^\ddagger)}{p_G(x_G^\ddagger)} \right)^\ddagger \cdot \left( \frac{1}{2\sigma^2} U_\cN(F(x_{G,i}^\ddagger)) - \log \left| \det \left( \frac{\partial F(x_{G,i}^\ddagger)}{\partial x_{G,i}^\ddagger} \right) \right| \right) \right]
\end{equation}

with $\ddagger$ the symbol used to denote the ``detach'' operator that makes the term constant with respect to gradient descent: $\bx_G^\ddagger \sim p_G^\ddagger$ is therefore a detached mini-batch of size $n$.

\begin{figure}[!b]
    \centering
    \begin{subfigure}{.30\textwidth}
        \includegraphics[width=117pt, height=50pt]{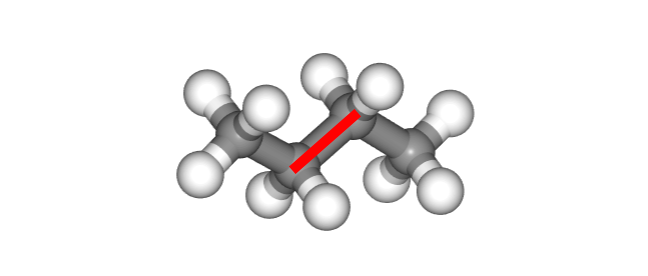}
        \vspace{10pt}
        \caption{A configuration of \texttt{Butane}}
        \label{fig:PP_butane_molecule}
    \end{subfigure}
    \begin{subfigure}{.68\textwidth}\hspace{12pt}
        \includegraphics[width=240pt, height=68pt]{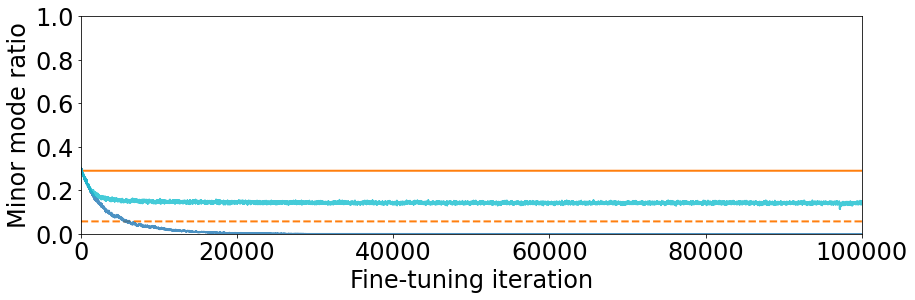}
        \caption{Percentage of generated samples $x_G \sim p_G$ in the minor modes.}
        \label{fig:PP_butane_lambda0_finetuning_ratios}
    \end{subfigure}
    \begin{subfigure}{.24\textwidth}\centering
        \includegraphics[width=94pt, height=80pt]{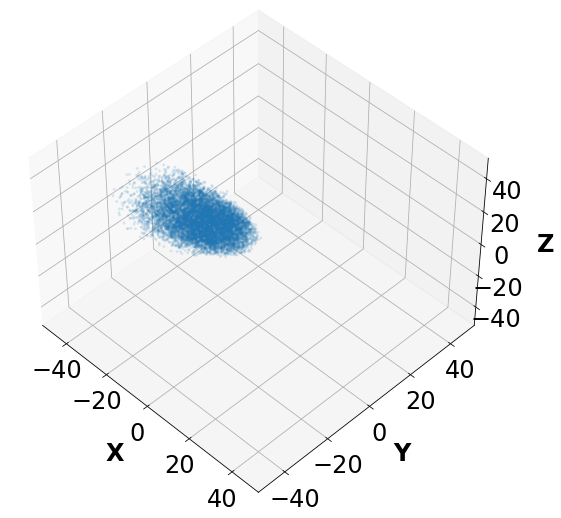}
        \caption{Centers of mass}
        \label{fig:PP_butane_lambda0_KLx_translations}
    \end{subfigure}
    \begin{subfigure}{.24\textwidth}\centering
        \includegraphics[width=94pt, height=68pt]{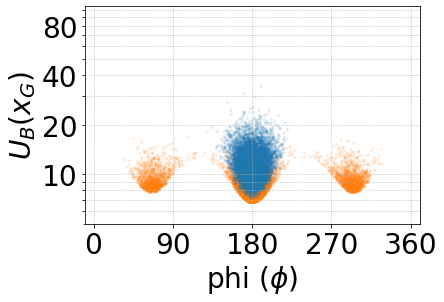}
        \caption{$U_B$ energies}
        \label{fig:PP_butane_lambda0_KLx_UBxG}
    \end{subfigure}
    \begin{subfigure}{.24\textwidth}\centering
        \includegraphics[width=94pt, height=68pt]{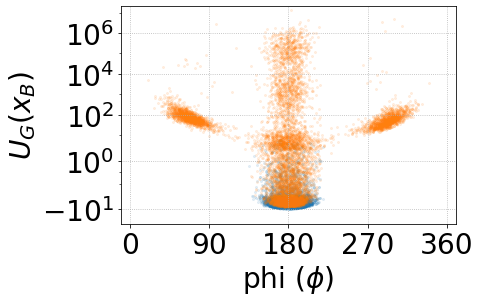}
        \caption{$U_G$ energies}
        \label{fig:PP_butane_lambda0_KLx_UGxB}
    \end{subfigure}
    \begin{subfigure}{.24\textwidth}\centering
        \includegraphics[width=80pt, height=68pt]{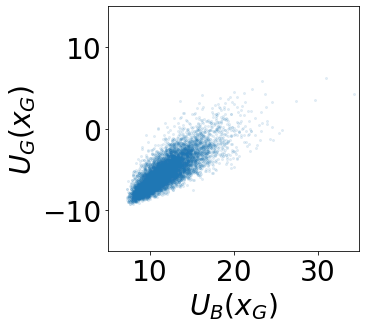}
        \caption{Correlations}
        \label{fig:PP_butane_lambda0_KLx_correlations}
    \end{subfigure}
    \begin{subfigure}{.24\textwidth}\centering
        \includegraphics[width=94pt, height=80pt]{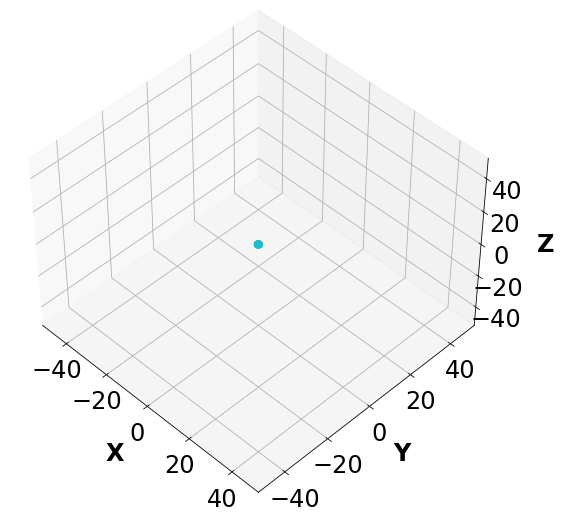}
        \caption{Centers of mass}
        \label{fig:PP_butane_lambda0_KLz_translations}
    \end{subfigure}
    \begin{subfigure}{.24\textwidth}\centering
        \includegraphics[width=94pt, height=68pt]{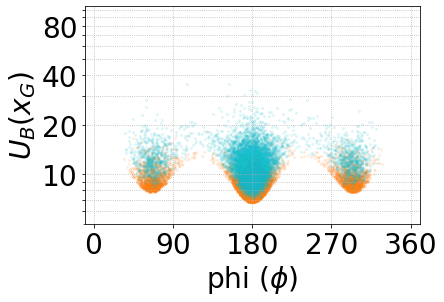}
        \caption{$U_B$ energies}
        \label{fig:PP_butane_lambda0_KLz_UBxG}
    \end{subfigure}
    \begin{subfigure}{.24\textwidth}\centering
        \includegraphics[width=94pt, height=68pt]{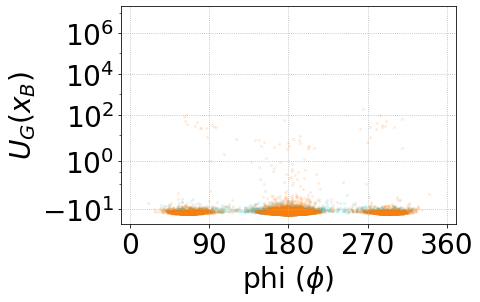}
        \caption{$U_G$ energies}
        \label{fig:PP_butane_lambda0_KLz_UGxB}
    \end{subfigure}
    \begin{subfigure}{.24\textwidth}\centering
        \includegraphics[width=80pt, height=68pt]{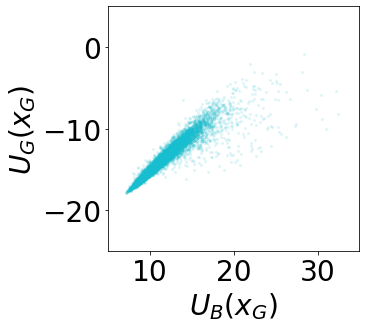}
        \caption{Correlations}
        \label{fig:PP_butane_lambda0_KLz_correlations}
    \end{subfigure}
    \caption{
        Results of two fine-tunings with $\cL_{\KL x}$ (second row) and $\cL_{\KL z}^\mathrm{df}$ (third row) after the same pre-training with $\cL_{\KL z}$ on \texttt{Butane}. In all sub-figures, data from $p_B$ (i.e. the dataset) is represented in \textcolor{PltOrange}{\ul{orange}}, fine-tuning results with $\cL_{\KL x}$ are represented in \textcolor{PltBlue}{\ul{blue}}, fine-tuning results with $\cL_{\KL z}^\mathrm{df}$ are represented in \textcolor{PltCyan}{\ul{cyan}}.
        \\- Figures~\ref{fig:PP_butane_lambda0_KLx_translations} and \ref{fig:PP_butane_lambda0_KLz_translations} represent the centers of mass of generated samples $x_G \sim p_G$.
        \\- Figures~\ref{fig:PP_butane_lambda0_KLx_UBxG} and \ref{fig:PP_butane_lambda0_KLz_UBxG} represent the potential energy $U_B$ of generated samples $x_G \sim p_G$ (in \textcolor{PltBlue}{\ul{blue}} or \textcolor{PltCyan}{\ul{cyan}}) vs. samples from the dataset $x_B \sim p_B$ (in \textcolor{PltOrange}{\ul{orange}}). Note that in both cases the energy of the hydrogens is minimized (either by the model or manually). These figures visualize whether or not $p_G \subset p_B$.
        \\- Figures~\ref{fig:PP_butane_lambda0_KLx_UGxB} and \ref{fig:PP_butane_lambda0_KLz_UGxB} represent the energy of generation $U_G$ of samples from the dataset $x_B \sim p_B$ according to each pre-trained model (in \textcolor{PltOrange}{\ul{orange}}) vs generated samples $x_G \sim p_G$ (in \textcolor{PltBlue}{\ul{blue}} or \textcolor{PltCyan}{\ul{cyan}}). These figures visualize whether or not $p_B \subset p_G$.
        \\- Figures~\ref{fig:PP_butane_lambda0_KLx_correlations} and \ref{fig:PP_butane_lambda0_KLz_correlations} represent the correlations between the energy of generation $U_G$ and the potential energy $U_B$ of generated samples $x_G \sim p_G$.
    }
    \label{fig:PP_butane_lambda0}
\end{figure}

This results in a loss that is significantly more stable than $\cL_{\KL x}$ and works perfectly on \texttt{Double well 12D} (data not shown). It also achieves good performance on more complex target distributions like that of \texttt{Butane} (figure~\ref{fig:PP_butane_lambda0}). The configurations of the butane molecule have three main modes that can easily be visualized when projecting onto the values of the dihedral angle $\phi$ of its carbon chain (in red, figure~\ref{fig:PP_butane_molecule}).

The potential energy function $U_B$ of physical systems in the absence of external fields is invariant by collective rotation and translation.
When the generative model is expressed in Cartesian coordinates and is not equivariant with respect to these external degrees of freedom, it is necessary to add a loss term that discourages translations and rotations, essentially acting as an alignment penalty. This penalty is weighted by a scalar denoted $\lambda_\mathrm{align}$. To showcase the different behaviors of $\cL_{\KL x}$ and $\cL_{\KL z}^\mathrm{df}$, a model is pre-trained with $\lambda_\mathrm{align} = 10$ and then fine-tuned with $\lambda_\mathrm{align} = 0$, essentially asking the generated density to expand infinitely in the translational degrees of freedom and to cover all possible rotations.

$\cL_{\KL x}$ makes $p_G$ continuously expand translationally (figure~\ref{fig:PP_butane_lambda0_KLx_translations}) but at the expense of losing the minor modes (figures~\ref{fig:PP_butane_lambda0_finetuning_ratios} and \ref{fig:PP_butane_lambda0_KLx_UBxG}), resulting in an explosion of the energy of generation $U_G$ (figure~\ref{fig:PP_butane_lambda0_KLx_UGxB}). $\cL_{\KL z}^\mathrm{df}$ on the other hand, does \textit{not} explore significantly (figure~\ref{fig:PP_butane_lambda0_KLz_translations}) but remains very stable by keeping all the modes (figure~\ref{fig:PP_butane_lambda0_KLz_UGxB}) and producing samples that stay at low energy levels (figures~\ref{fig:PP_butane_lambda0_KLz_UBxG} and \ref{fig:PP_butane_lambda0_KLz_correlations}).

While these results describe the extreme case of degrees of freedom distributed uniformly over $\mathbb{R}$, they exemplify the importance of removing unnecessary degrees of freedom for better performance, especially those whose broad distribution considerably expands the support of the target density. For translations and rotations, this can be achieved by always using $\lambda_\mathrm{align} > 0$. Of note, the degrees of freedom of hydrogen atoms (which are permutation invariant within groups like $-CH3$ for example) are also ignored here. The model is only asked to generate the positions of the carbon atoms, and another module places the hydrogen atoms deterministically near their energy minimum. This introduces a bias and changes the target ratio between the modes (from the solid to the dashed line in figure~\ref{fig:PP_butane_lambda0_finetuning_ratios}, see appendix~\ref{sec:appendix:mode_ratio_change}) but does not explain why $\cL_{\KL z}^\mathrm{df}$ does not converge to the expected (``dashed'') ratio. $\cL_{\KL z}^\mathrm{df}$ is also shown to be unstable on more complex datasets (i.e. \texttt{Dialanine}, figure~\ref{fig:PP_dialanine_lambda10_finetuning_UBs}) and a better loss is developed in section~\ref{sec:PL2} to counteract this problem.

Since divergences are not symmetric, one might also wonder what happens when swapping the two distributions within the $\KL$ divergences, but an important result from the literature~\cite{Papamakarios2019NFsReview} already shows that the minimizations of $\KL(q_\cN||q_F)$ and $\KL(p_G||p_B)$ are equivalent, as well as the minimizations of $\KL(p_B||p_G)$ and $\KL(q_F||q_\cN)$. $\KL(p_B||p_G)$ is known to often lead to mode-drop in $x$-space~\cite{Murphy2012MachineLearning}, whereas $\KL(q_\cN||q_F)$ tends to avoid that behavior (in our case, it may cause mode collapse in $z$-space but this is not an impediment since the Gaussian target distribution has only one mode). Note also that combining both data-free losses ($\cL_{\KL z}^\mathrm{df}$ and $\cL_{\KL x}$) is not sufficient to get proper ratios since $\cL_{\KL x}$ tend to dominate and the fine-tuning still results in a mode-collapse.

\paragraph{Optimization pitfalls due to discretization over minibatches.}
In Appendix \ref{sec:appendix:pitfalls}, we show that in general the optimization of Kullback-Leibler divergences with respect to a distribution suffers from severe issues when discretized over minibatches without proper normalization. This is due to the fact that the properties of $\KL$ heavily rely on a global unit mass constraint (for Gibbs inequality to hold), which hinders its estimation or optimization in practice.
We show how to build more suitable estimators of the gradient of the Kullback-Leibler divergence, as well as how to minimize their variance via a \emph{stabilizing trick}.

%%%%%%%%%%%%%%%%%%%%%%%%%%%%%%%%%%%%%%%%%%%%%%%%%%%%%%%%%%%%%%%%%%%%%%%%%%%%%%%%%%%%%%%%%%%%%%%%%%%
%%%%%%%%%%%%%%%%%%%%%%%%%%%%%%%%%%%%%%%%%%%%%%%%%%%%%%%%%%%%%%%%%%%%%%%%%%%%%%%%%%%%%%%%%%%%%%%%%%%
\section{Desirable Properties for a Loss Function}
\label{sec:design}

\subsection{Estimator variance as a loss}
\label{subsec:design:estimator_variance_as_loss}

With normalizing flows, one can compute exactly the probability with which one generates any given point. As a consequence, one can correct the sampler based on the trained generator with importance sampling, i.e. by associating each sample $x$ with a weight $\frac{p_B(x)}{p_G(x)}$. Expectations are then taken with respect to $\frac{p_B}{p_G} p_G$, which \emph{exactly} matches the target $p_B$, regardless of $p_G$ (provided that it has positive density everywhere $p_B$ does). However, if importance sampling weights are closer to 1, the produced distribution will converge faster towards $p_B$, that is, fewer samples will need to be generated. The question here is how to design a loss to train $p_G$ in such a context where the reweighted output distribution is always perfect.

An important application of our generator $G$ is often to estimate integral quantities of the form $\E_{p_B}[f]$ for some given function $f$. For instance, a classic use case in practice is to compute the free energy difference $\Delta F_{BC}$ between the state being sampled (with energy $U_B$) and an alternate state (with energy $U_C$). Then:
\begin{equation}
    e^{-\beta \Delta F_{BC}} := \E_{x \sim p_B} [f] \qquad \text{ with } \qquad f(x) = e^{-\beta(U_C(x)-U_B(x))}
\end{equation}

Let us denote by $Q$ the true value of the quantity to estimate:
\begin{equation}
    Q := \E_{p_B}[ f ] := \int_{x \in \cX} p_B(x) f(x) dx = \E_{p_G}\left[ \frac{p_B}{p_G} \cdot f \right]
\end{equation}

The latter equality holds under the assumption that $p_G$ is never 0 where $p_B$ is not. For any $p_G$, the following quantity $\hQ$ is an unbiased estimator of $Q$:
\begin{equation}
    \hQ := \frac{1}{n} \sum_{x_i \in m} f(x_i) \frac{p_B(x_i)}{p_G(x_i)}
\end{equation}

where $m = (x_1, \dots, x_N)$ is a large set of points sampled according to $p_G$. That is, when averaging over all possible mini-batches, $\hQ$ becomes $Q$ (i.e. $\E_m[\hQ] = Q$). Yet, for some distributions $p_G$, the estimate $\hQ$ may converge faster than others, in terms of number of samples required to reach a given accuracy. The quality of a generator $p_G$ can thus be quantified through the expected error when estimating $Q$ with $n$ points. This can be shown to be proportional to the variance of $\hQ$, which can then be turned into a training loss (see appendix~\ref{appendix:proof_variance_loss} for a proof):
\begin{equation}
    \cL_f(p_G) = \E_{x \sim p_G}\left[ \frac{p_B^2(x)}{p_G^2(x)} \cdot f^2(x) \right]
\end{equation}

If the function $f$ is not fixed and can be any bounded function over the space $\cX$ of points $x$, then one can deduce the following optimization criterion:
\begin{equation}
    \label{eq:design:L}
    \cL(p_G) =\E_{x \sim p_G}\left[ \frac{p_B^2(x)}{p_G^2(x)} \right] = \E_{x \sim p_B}\left[ \frac{p_B(x)}{p_G(x)} \right] = e^{RN_2(p_B||p_G)} = \var_{x \sim p_G}\left[ \frac{p_B}{p_G} \right] + 1
\end{equation}

where $RN_2(p_B||p_G)$ is the Rényi divergence of order 2. This formula looks very similar to the $\KL$ divergence, without the $\log$, thus penalizing high ratios $\frac{p_B}{p_G}$ more strongly. In practice, one knows how to compute $\tpB(x) := \mathcal{Z}_B p_B(x)$ but not $p_B(x)$ directly. Fortunately, a model $p_G$ trained with $\cL$ will yield by definition a good estimator of $\mathcal{Z}_B =  \E_{p_B}\left[ \mathcal{Z}_B \right] = \E_{p_G}\left[ \frac{\tpB}{p_G} \right]$. 

Another justification for this loss is that one aims to find $p_G \propto \tpB$, and therefore to make the ratio $\frac{\tpB}{p_G}$ constant over $\cX$. Without knowing the value of the target constant, this can still be achieved by minimizing the variance of the ratio over $\cX$, which is precisely the loss $\cL$.

Thus we arrive at $\cL(p_G) = \var_{x \sim p_G}\left[ \frac{p_B}{p_G} \right]$ as a principled loss to minimize the variance of estimators of expectations over the Boltzmann distribution.

\subsection{Practical Recommendations}
\label{subsec:design:recommendations}

Beyond the theoretical points considered in section~\ref{subsec:design:estimator_variance_as_loss}, there are a few practical considerations that need to be addressed.

\textbf{Degrees of freedom:}
\begin{itemize}[leftmargin=20pt, noitemsep, topsep=0pt]
    \item[-] As illustrated in section~\ref{sec:KLs}, avoiding unnecessary symmetries within the target distribution is often beneficial to ease the training. Hydrogen atoms for instance are permutation invariant within $-CH3$ groups and thus multiply by 6 the total number of modes for each group. Since the position of hydrogen atoms is often irrelevant for downstream applications they can often be ignored. In this work we choose the simplest method which consist in placing the hydrogens deterministically near their energy minimum at the cost of intruducing a bias that changes the ratio between modes. Better options exist such as adjusting $U_B$ (to encourage having only one permutation possible), or placing hydrogen atoms stochastically but with a model that does not care about mode collapse.
    \item[-] More importantly, extremely flat degrees of freedom should be removed if possible. When it comes to translations and rotations, several approaches are available. One could add an alignment penalty to the potential energy $U_B$ (as described in section~\ref{sec:KLs}), but it is also possible to generate configurations in internal coordinates directly (thereby removing 6 degrees of freedom).
\end{itemize}

\textbf{Numerical instabilities:}
\begin{itemize}[leftmargin=20pt, noitemsep, topsep=0pt]
    \item[-] The loss $\cL_{\KL z}^\mathrm{df}$ may suffer from training instabilities due to the use of importance sampling weights that have a high variance and therefore often focuses most of the gradient onto just a few points of each mini-batch. Such weights should be avoided if possible during the design of new loss functions.
    \item[-] The potential energy term $U_B$ is also at risk of introducing training instabilities since it can be very sensitive to small changes in the position of the atoms. The strategy followed in this work is to cap each term of the energy function individually, so that their gradient never exceeds a given threshold. This approach is much more fine-grained than using a global capping, directly on $U_B$.
\end{itemize}

\textbf{Minimizing vs. maximizing the energy terms:}
\begin{itemize}[leftmargin=20pt, noitemsep, topsep=0pt]
    \item[-] The term $U_B$ should probably never be increased explicitly through gradient descent (which is equivalent to saying that $p_B$ should never be decreased). Although some training objectives that do this may \textit{seem} to be principled in the context of an integral over the whole space, they usually fail once converted into loss functions used on discrete mini-batches.
    \item[-] In the same spirit, it is often a preferable to avoid decreasing $p_G$ directly. Indeed, decreasing $p_G$ at a given point implies moving the mass somewhere else, but since the direction where to move this mass is not specified, it could go anywhere without actually getting any closer to $p_B$. Since $p_G$ is a probability distribution, increasing it anywhere implies that some other region of the space will become less probable to compensate (i.e. $p_G$ cannot increase everywhere). In the case where $p_G$ is never decreased explicitly (maybe by masking the troublesome points) the training is much smoother since the probability mass is always pushed where it is most needed.
\end{itemize}

%%%%%%%%%%%%%%%%%%%%%%%%%%%%%%%%%%%%%%%%%%%%%%%%%%%%%%%%%%%%%%%%%%%%%%%%%%%%%%%%%%%%%%%%%%%%%%%%%%%
%%%%%%%%%%%%%%%%%%%%%%%%%%%%%%%%%%%%%%%%%%%%%%%%%%%%%%%%%%%%%%%%%%%%%%%%%%%%%%%%%%%%%%%%%%%%%%%%%%%
\section{A data-free \texorpdfstring{$L^2$}{L2} loss}
\label{sec:PL2}

\begin{figure}[!b]
    \centering
    \begin{subfigure}{.30\textwidth}\centering
        \includegraphics[width=80pt, height=68pt]{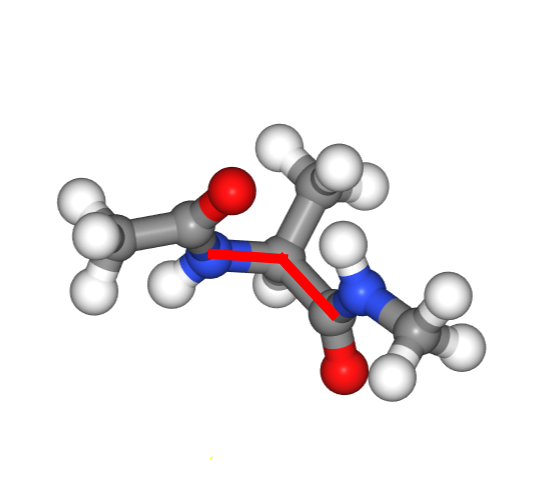}
        \caption{A configuration of dialanine}
        \label{fig:PP_dialanine_molecule}
    \end{subfigure}
    \begin{subfigure}{.68\textwidth}\hspace{12pt}
        \includegraphics[width=240pt, height=68pt]{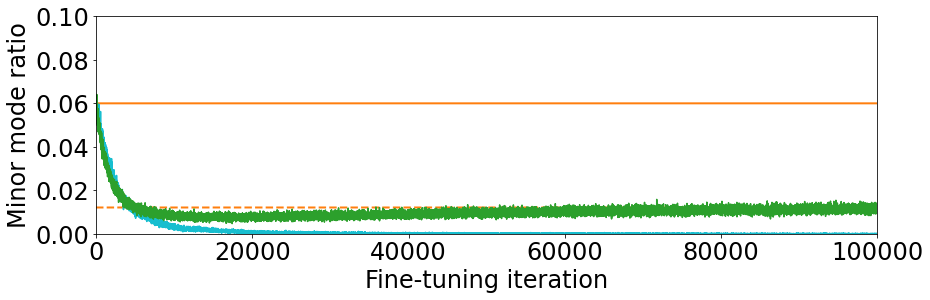}
        \caption{Percentage of generated samples $x_G \sim p_G$ in the minor modes.}
        \label{fig:PP_dialanine_lambda10_finetuning_ratios}
    \end{subfigure}
    \begin{subfigure}{.98\textwidth}\centering
        \includegraphics[width=90pt, height=75pt]{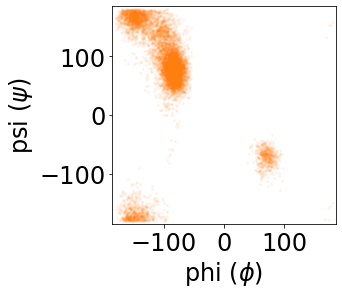}\hspace{7pt}
        \includegraphics[width=90pt, height=75pt]{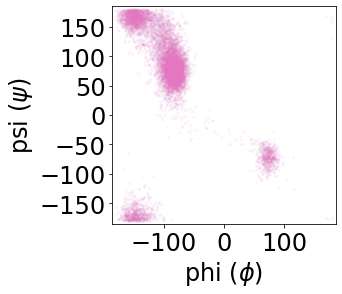}\hspace{7pt}
        \includegraphics[width=90pt, height=75pt]{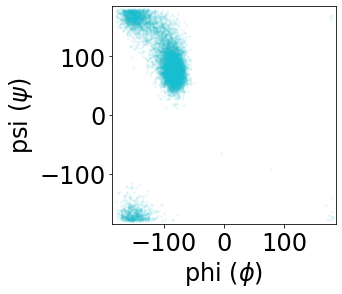}\hspace{7pt}
        \includegraphics[width=90pt, height=75pt]{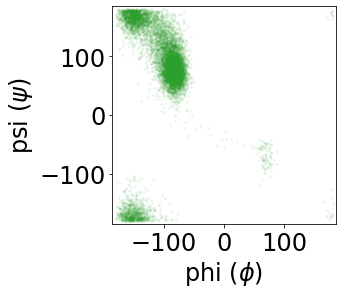}
        \caption{Projections on $(\phi, \psi$) dihedral angles. Ground truth target (\textcolor{PltOrange}{\ul{orange}}), model after data-dependent pre-training (\textcolor{PltPink}{\ul{pink}}), models fine-tuned with $\cL_{\KL z}^\mathrm{df}$ (\textcolor{PltCyan}{\ul{cyan}}) and $\cL_{L^2_+}$ (\textcolor{PltGreen}{\ul{green}})}
        \label{fig:PP_dialanine_lambda10_PL2cache_ramachandrans}
    \end{subfigure}
    \begin{subfigure}{.250\textwidth}\centering
        \includegraphics[width=97pt, height=68pt]{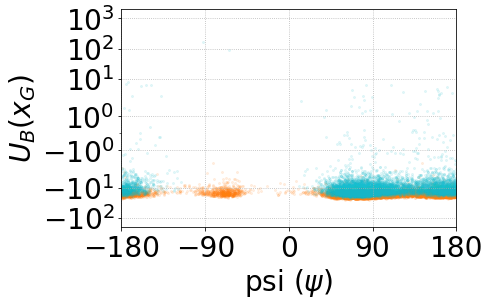}
        \caption{$U_B$ energies with $\cL_{\KL z}^\mathrm{df}$}
        \label{fig:PP_dialanine_lambda10_KLz_UBxG}
    \end{subfigure}
    \begin{subfigure}{.240\textwidth}\centering
        \includegraphics[width=97pt, height=68pt]{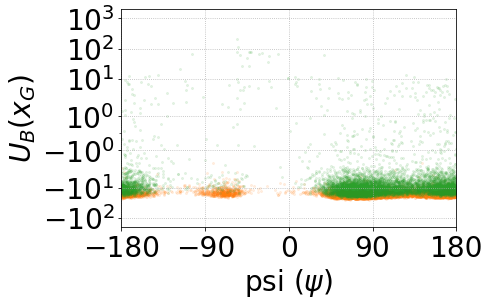}
        \caption{$U_B$ energies with $\cL_{L^2_+}$}
        \label{fig:PP_dialanine_lambda10_PL2cache_UBxG}
    \end{subfigure}
    \begin{subfigure}{.250\textwidth}\centering
        \includegraphics[width=97pt, height=68pt]{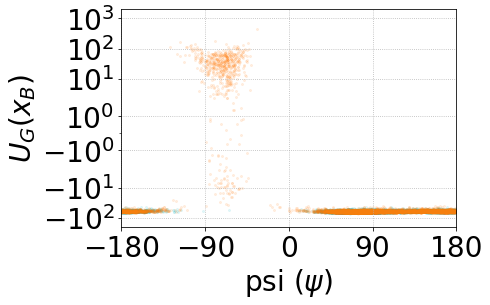}
        \caption{$U_G$ energies with $\cL_{\KL z}^\mathrm{df}$}
        \label{fig:PP_dialanine_lambda10_KLz_UGxB}
    \end{subfigure}
    \begin{subfigure}{.240\textwidth}\centering
        \includegraphics[width=97pt, height=68pt]{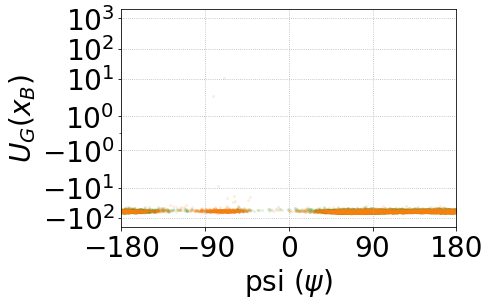}
        \caption{$U_G$ energies with $\cL_{L^2_+}$}
        \label{fig:PP_dialanine_lambda10_PL2cache_UGxB}
    \end{subfigure}
    \begin{subfigure}{.30\textwidth}\hspace{20pt}
        \includegraphics[width=80pt, height=68pt]{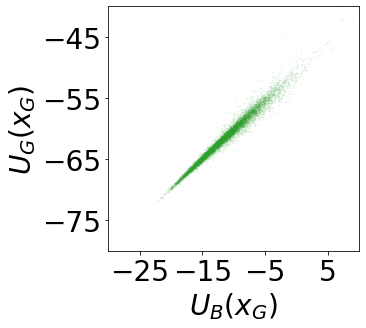}
        \caption{Correlations}
        \label{fig:PP_dialanine_lambda10_PL2cache_correlations}
    \end{subfigure}
    \begin{subfigure}{.68\textwidth}\hspace{12pt}
        \includegraphics[width=240pt, height=68pt]{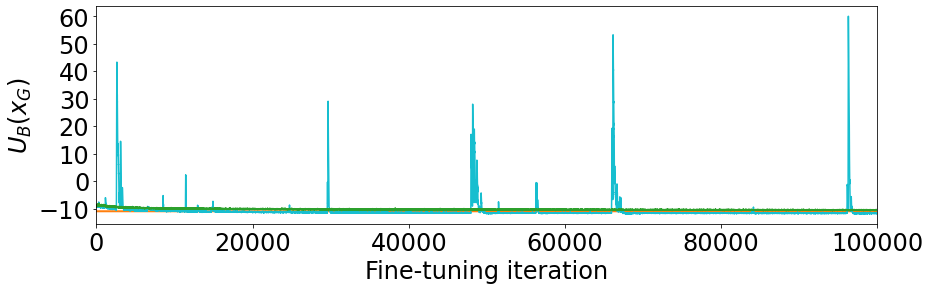}
        \caption{$U_B$ energy of generated samples $x_G \sim p_G$ during fine-tuning.}
        \label{fig:PP_dialanine_lambda10_finetuning_UBs}
    \end{subfigure}
    \caption{
        Results of two fine-tunings (with $\cL_{\KL z}^\mathrm{df}$ and $\cL_{L^2_+}$) after the same pre-training with $\cL_{\KL z}$ on \texttt{Dialanine}. In all sub-figures, data from $p_B$ (i.e. the dataset) is colored in \textcolor{PltOrange}{\ul{orange}}, pre-training results are colored in \textcolor{PltPink}{\ul{pink}}, fine-tuning results with $\cL_{\KL z}^\mathrm{df}$ are colored in \textcolor{PltCyan}{\ul{cyan}}, fine-tuning results with $\cL_{L^2_+}$ are colored in \textcolor{PltGreen}{\ul{green}}.
        \\- Figure~\ref{fig:PP_dialanine_lambda10_finetuning_ratios} represents the percentage of generated samples $x_G \sim p_G$ in the minor modes during fine-tuning. The solid \textcolor{PltOrange}{\ul{orange}} line corresponds to the ``real'' ratio from $p_B$, whereas the dashed \textcolor{PltOrange}{\ul{orange}} line corresponds to the same ratio from $p_B$ but with the energy minimized with repect to hydrogen atom coordinates (appendix~\ref{sec:appendix:mode_ratio_change}).
        \\- Figure~\ref{fig:PP_dialanine_lambda10_PL2cache_ramachandrans} contains 2D projections of the ground truth dataset and generated data.
        \\- Figures~\ref{fig:PP_dialanine_lambda10_KLz_UBxG} and \ref{fig:PP_dialanine_lambda10_PL2cache_UBxG} represent the potential energy $U_B$ of generated samples $x_G \sim p_G$ (in \textcolor{PltCyan}{\ul{cyan}} or \textcolor{PltGreen}{\ul{green}}) vs. samples from the dataset $x_B \sim p_B$ (in \textcolor{PltOrange}{\ul{orange}}). Note that in both cases the energy of the hydrogens is minimized (either by the model or manually). These figures visualize whether or not $p_G \subset p_B$.
        \\- Figures~\ref{fig:PP_dialanine_lambda10_KLz_UGxB} and \ref{fig:PP_dialanine_lambda10_PL2cache_UGxB} represent the energy of generation $U_G$ of samples from the dataset $x_B \sim p_B$ (in \textcolor{PltOrange}{\ul{orange}}) vs generated samples $x_G \sim p_G$ (in \textcolor{PltCyan}{\ul{cyan}} or \textcolor{PltGreen}{\ul{green}}). These figures visualize whether or not $p_B \subset p_G$.
        \\- Figures~\ref{fig:PP_dialanine_lambda10_PL2cache_correlations} represents the correlations between the energy of generation $U_G$ and the potential energy $U_B$ of generated samples $x_G \sim p_G$.
        \\- Figure~\ref{fig:PP_dialanine_lambda10_finetuning_UBs} represents the potential energy $U_B$ of generated samples $x_G \sim p_G$ during fine-tuning. Note the instability of $\cL_{\KL z}^\mathrm{df}$ compared to the stability of $\cL_{L^2_+}$.
    }
    \label{fig:PP_dialanine_lambda10}
\end{figure}

Building on $\var_{x \sim p_G}\left[ \frac{p_B}{p_G} \right]$ (from equation~\ref{eq:design:L}), we replace ratios $\frac{p_B(x)}{p_G(x)}$ with log-ratios
\begin{equation}
    r(x) = \log\frac{p_B(x)}{p_G(x)}
\end{equation}
for numerical reasons, as normalizing flows actually compute $\log$-probabilities and the exponentiation leads to instability.
We also note that:
\begin{equation}
    \DDS\var_{p_G}[r] = \E_{p_G}\left[ \left(r - \E_{p_G}[r] \right)^2 \right]
\end{equation}

This formulation with differences between log-ratios has the advantage of making $\mathcal{Z}_B$ cancel out from the computations in practice. To avoid decreasing $p_G(x)$ explicitly at any point $x$, as mentioned in Section~\ref{subsec:design:recommendations}, we modify the loss as follows by masking $(r - K)$. As a consequence, $r$ (and therefore $U_G$) can only be minimized (whereas $U_B(x_G^\ddagger)$ is not differentiated with respect to $\theta$). The masked $L^2$ loss with detached means is therefore defined as:

\begin{equation}
    \label{eq:KLs:PL2_loss}
    \cL_{L^2_+}(\bx_G^\ddagger) = \sum_{i=1}^n \left[ \frac{1}{n} \cdot \left[ \left(r(x_{G, i}^\ddagger) -  K^\ddagger \right)_+^2\right] \right]
\end{equation}

where $a_+^2 = a^2$ if $a>0$ and 0 otherwise, and where $K^\ddagger = \left[ \sum_{j=1}^n \left[ \frac{1}{n} \cdot r(x_{G, j}^\ddagger) \right] \right]^\ddagger$ is not differentiated (so as to ensure that it is never increased). Note that in the continuous limit: $\E_{p_G}[r] = -\KL(p_G||p_B) \leqslant 0$.

One can prove that, in spite of the non-differentiation of $K^\ddagger$, a pseudo-gradient descent on this loss \emph{will} converge towards $p_B$, provided the model is expressive enough and that the initial $p_G$ is non-zero on the support of $p_B$, for an adequate choice of inner product (appendix~\ref{sec:appendix:L2_loss_convergence}).

This loss has common features with log-variance loss of Richter et al.~\cite{Richter2020}, yet the mask applied in the present loss is critical for stability, just as well as detaching $K$ (see the ablation study in appendix~\ref{sec:appendix:detach_removal}).

The conformational distribution of dialanine is often projected onto its two main dihedral angles $\phi$ and $\psi$ for visualization (figures~\ref{fig:PP_dialanine_molecule} and \ref{fig:PP_dialanine_lambda10_PL2cache_ramachandrans}). The ``real'' distribution $p_B$ has about $\approx 6\%$ of its mass in the minor mode (the one where $\phi > 0$) but when taking into account the minimization of the energy of hydrogen atoms, this ratio drops to $\approx 1.21\%$ (dashed line in figure~\ref{fig:PP_dialanine_lambda10_finetuning_ratios}, see appendix~\ref{sec:appendix:mode_ratio_change}). This means that the result of the pre-training with the data-dependent $\cL_{\KL z}$ produces a ratio ($\approx 6\%$, figure~\ref{fig:PP_dialanine_lambda10_PL2cache_ramachandrans}) that is different from the one expected at the end of the fine-tuning ($\approx 1.21\%$). It is clear that $\cL_{\KL z}^\mathrm{df}$ completely loses the minor mode (figures \ref{fig:PP_dialanine_lambda10_PL2cache_ramachandrans} and \ref{fig:PP_dialanine_lambda10_KLz_UGxB}) whereas $\cL_{L^2_+}$ does not (figures~\ref{fig:PP_dialanine_lambda10_PL2cache_ramachandrans} and \ref{fig:PP_dialanine_lambda10_PL2cache_UGxB}) and converges to the expected correct ratio of $\approx 1.21\%$. The bias induced by the deterministic placement of the hydrogen atoms does not change the main point that $\cL_{L^2_+}$ converged to the ratio it was supposed to produce. Another thing to notice is that $\cL_{\KL z}^\mathrm{df}$ has unstable $U_B$ energies during fine-tuning whereas the $\cL_{L^2_+}$ does not (figure~\ref{fig:PP_dialanine_lambda10_finetuning_UBs}).
In addition to those clear qualitative improvements, and unlike $\cL_{\KL z}^\mathrm{df}$, $\cL_{L^2_+}$ does not rely on numerically unstable importance sampling weights.

Note that the accuracy on this test is limited by the choice of generating deterministic hydrogen atom positions: this can be lifted by using a conditional normalizing flow to generate a Boltzmann distribution of hydrogen atom positions conditioned on the set of heavy atom positions generated by the main model.

%%%%%%%%%%%%%%%%%%%%%%%%%%%%%%%%%%%%%%%%%%%%%%%%%%%%%%%%%%%%%%%%%%%%%%%%%%%%%%%%%%%%%%%%%%%%%%%%%%%
%%%%%%%%%%%%%%%%%%%%%%%%%%%%%%%%%%%%%%%%%%%%%%%%%%%%%%%%%%%%%%%%%%%%%%%%%%%%%%%%%%%%%%%%%%%%%%%%%%%
\section{Technical details}
\label{sec:methods}

\paragraph{Data and pretraining.}
For all datasets (i.e. \texttt{Double Well 12D}, \texttt{Butane} and \texttt{Dialanine}), the data has been generated by Metropolis-Hastings simulations with Parallel Tempering~\cite{Swendsen1986ReplicaExchange, Sugita1999ReplicaExchange, Earl2005ParallelTempering}).
The potential energy $U_B$ of the molecules of butane and dialanine is evaluated according to the \href{https://www.charmm.org/archive/charmm/resources/charmm-force-fields}{CHARMM36m force field}~\cite{Huang2016}.
The energy function used in the simulations also uses an alignment penalty with $\lambda_\mathrm{align} = 10$.
The data-dependent pre-training is performed with $L_{\KL z}$. The number of iterations was a 10th of the one used for fine-tunings (i.e. 10000 iterations). 

\paragraph{Alignment Penalty.}
The alignment penalty is an L2 distance between generated coordinates and their image after a roto-translational alignment to some reference. The alignment may only be partial since we cap the maximum allowed rotation by $\pi/3$ to ease the training.

\paragraph{Model Architecture.}
Only two model architectures are used. One for \texttt{Double Well 12D} and one for molecular datasets (i.e. \texttt{Butane} and \texttt{Dialanine}). The architecture used in \texttt{Double Well 12D} experiments is a simple stack of $8 \times 4$ Coupling Blocks (as described in section~\cite{Dinh2016RealNVP}). Every Coupling Block uses an internal feed-forward sub-network $M$ composed of two layers with an internal feature size of 64 separated by a CELU non-linearity~\cite{Barron2017CELU}. The architecture used in \texttt{Butane} and \texttt{Dialanine} experiments is quite similar except for two changes:
\begin{itemize}[leftmargin=20pt, noitemsep, topsep=0pt]
    \item[-] The stack is made deeper ($24 \times 4$ Coupling Blocks) and wider (internal sub-networks $M$ have a layer size of 256),
    \item[-] and an additional feed-forward network is used to generate the position of the hydrogen atoms. It has 3 layers separated by CELUs and a hidden size of 512.
\end{itemize}

\paragraph{Error estimation.}
No error bars are provided, but empirically, each experiment proved to be entirely reproducible over dozens of runs.

\paragraph{Resources.}
Every experiment has been performed on a single machine with two GPUs GeForce RTX\texttrademark{} 2070. Each experiment on \texttt{Double Well 12D} takes about 40m, whereas the experiments on \texttt{Butane} and \texttt{Dialanine} take between 7 to 10 hours.

%%%%%%%%%%%%%%%%%%%%%%%%%%%%%%%%%%%%%%%%%%%%%%%%%%%%%%%%%%%%%%%%%%%%%%%%%%%%%%%%%%%%%%%%%%%%%%%%%%%
%%%%%%%%%%%%%%%%%%%%%%%%%%%%%%%%%%%%%%%%%%%%%%%%%%%%%%%%%%%%%%%%%%%%%%%%%%%%%%%%%%%%%%%%%%%%%%%%%%%
\section{Conclusion and Perspectives}
\label{sec:conclusion}

In this contribution, we have explored the conditions necessary for training or refining flow-based models based on an explicitly known target density, rather than pre-determined samples from a Markov-chain simulation.
We have found that several losses that may seem appropriate in theory lead to numerical failures in a discrete setting. In particular, we have documented a major instability issue when optimizing the $\KL$ divergence $\KL(p_G||p_B)$ between the generated and target distributions. We note that loss functions whose minimization amounts to decreasing the probability of a sample point (lowering either $p_G$ or $p_B$) push the model to spread local mass in improbable directions, resulting in instability.
Based on an estimator variance minimization approach, we have derived a stable data-free loss based on $L^2$ distances between log-distributions, with the important condition that a mask must be applied to follow the criterion stated above.
This loss is the first one to exhibit stable data-free optimization on the dialanine molecule task.

While this allows for stable optimization of a correctly trained model, lifting the requirement for complete reference data will require a training protocol able to explore the target space to discover new modes. 
We envision two families of approaches to that effect:
\begin{itemize}[leftmargin=20pt, noitemsep, topsep=0pt]
    \item[-] Keeping the current paradigm of a generator fully trained on a single system, training could be initiated based on a limited and/or biased set of data, for example from high-temperature simulations, then extended using the properties of normalizing flows themselves~\cite{Dibak2020TemperatureSteerableFlows, Dibak2021TemperatureSteerableFlows}, enhanced-sampling simulations\cite{heninSampling}, or hybrid approaches~\cite{Gabrie2022AdaptiveMCwithNFs}.
    \item[-] Alternately, the cost of complete training for every new target could be reduced by transferring information between systems using curriculum learning. In the case of molecular targets, this would require a generalizing model, e.g. one based on graph convolutions~\cite{Velikovic2017GATs, Schutt2017Schnet, Liu2019GraphNFs}.
\end{itemize}

The novel masked $L^2_+$ loss has demonstrated remarkable stability on the \texttt{Dialanine} test case, which is a good benchmark for small molecules of pharmacological interest, and a smaller proof of concept for proteins.
It remains to be seen how it will scale to larger systems, yet previous work has shown the normalizing flow approach to scale to larger molecules in presence of a training dataset~\cite{Noe2019}.

%%%%%%%%%%%%%%%%%%%%%%%%%%%%%%%%%%%%%%%%%%%%%%%%%%%%%%%%%%%%%%%%%%%%%%%%%%%%%%%%%%%%%%%%%%%%%%%%%%%
%%%%%%%%%%%%%%%%%%%%%%%%%%%%%%%%%%%%%%%%%%%%%%%%%%%%%%%%%%%%%%%%%%%%%%%%%%%%%%%%%%%%%%%%%%%%%%%%%%%
% Acknowledgments  % todo
\begin{ack}
Funding to LF was provided by Inria through IPL HPC-BigData. We are grateful to Bruno Raffin for leading the consortium that created and supported this project. 
We also thank Victor Berger and Cyril Furtlehner for fruitful discussions.
\end{ack}

%%%%%%%%%%%%%%%%%%%%%%%%%%%%%%%%%%%%%%%%%%%%%%%%%%%%%%%%%%%%%%%%%%%%%%%%%%%%%%%%%%%%%%%%%%%%%%%%%%%
%%%%%%%%%%%%%%%%%%%%%%%%%%%%%%%%%%%%%%%%%%%%%%%%%%%%%%%%%%%%%%%%%%%%%%%%%%%%%%%%%%%%%%%%%%%%%%%%%%%
% References
{
    \small
    \bibliographystyle{unsrtnat}
    \bibliography{refs}
}

\newpage

\newcommand{\cM}{\mathcal{M}}
\newcommand{\tp}{\tilde{p}}

\appendix

\section{Derivation of \texorpdfstring{$\KL(q_F||q_\cN)$}{KL(qF||qN)}}
\label{sec:appendix:KLz}

{%\small
    \begin{subequations}
        \label{eq:KLz:dvt}
        \begin{align}
            \KL(q_F||& q_\cN) = \int q_F(z) \log \frac{q_F(z)}{q_\cN(z)} \ dz \label{subeq:KLz:dvt_a} \\
            & = \int q_F(z) \log\cZ_\cN \ dz + \int q_F(z) \log \frac{q_F(z)}{\tilde{q}_\cN(z)} \ dz \label{subeq:KLz:dvt_b} \\
            & = \log\cZ_\cN + \int q_F(z) \log \frac{q_F(z)}{\tilde{q}_\cN(z)} \ dz \label{subeq:KLz:dvt_c} \\
            & = \log\cZ_\cN + \int p_B(x) \log \frac{p_B(x) \cdot \left| \det \left( \frac{\partial F(x)}{\partial x} \right) \right|^{-1}}{\tilde{q}_\cN(F(x))} \ dx \label{subeq:KLz:dvt_d} \\
            & = \log\cZ_\cN - S_B + \int p_B(x) \log \frac{\left| \det \left( \frac{\partial F(x)}{\partial x} \right) \right|^{-1}}{\tilde{q}_\cN(F(x))} \ dx \label{subeq:KLz:dvt_e} \\
            & = \log\cZ_\cN - S_B + \int p_B(x) \log \frac{\left| \det \left( \frac{\partial F(x)}{\partial x} \right) \right|^{-1}}{e^{-\frac{1}{2\sigma^2} U_\cN(F(x))}} \ dx \label{subeq:KLz:dvt_f} \\
            & = \log\cZ_\cN - S_B + \underset{x_B \sim p_B}{\bbE} \left[ \frac{1}{2\sigma^2} U_\cN(F(x_B)) + \log \left| \det \left( \frac{\partial F(x_B)}{\partial x_B} \right) \right|^{-1} \right] \label{subeq:KLz:dvt_g} \\
            & = \log\cZ_\cN - S_B + \underset{x_B \sim p_B}{\bbE} \left[ \frac{1}{2\sigma^2} U_\cN(F(x_B)) - \log \left| \det \left( \frac{\partial F(x_B)}{\partial x_B} \right) \right| \right] \label{subeq:KLz:dvt_h}
        \end{align}
    \end{subequations}
}

with:
\begin{itemize}[noitemsep, topsep=0mm]
    \item[-] (\ref{subeq:KLz:dvt_a}) by definition of the $\KL$ divergence
    \item[-] (\ref{subeq:KLz:dvt_b}) by using $q_\cN = \frac{1}{\cZ_\cN} \tilde{q}_\cN$
    \item[-] (\ref{subeq:KLz:dvt_c}) by using $\int q_F(z) dz = 1$ (probabilities sum to one)
    \item[-] (\ref{subeq:KLz:dvt_d}) by substitution of $q_F(z)$ by the change of variable formula:
    \begin{equation}
        \label{eq:UF:change_of_var}
        \begin{aligned}
            q_F(z) \ dz &= p_B(F^{-1}(z)) \cdot \left| \det \left( \frac{\partial F^{-1}(z)}{\partial z} \right) \right| \ dz \\
            &= p_B(x) \cdot \left| \det \left( \frac{\partial F(x)}{\partial x} \right) \right|^{-1} \ dz \\
            &= p_B(x) \ dx
        \end{aligned}
    \end{equation}
    \item[-] (\ref{subeq:KLz:dvt_e}) by definition of the entropy: $S_B = S(p_B) = - \int p_B(x) \log p_B(x) dx$
    \item[-] (\ref{subeq:KLz:dvt_f}) by using: $\tilde{q}_\cN(z) = e^{-\frac{1}{2\sigma^2} U_\cN(x)}$
    \item[-] (\ref{subeq:KLz:dvt_g}) by definition of expectation: $\underset{x_B \sim p_B}{\bbE} \big[ f(x_B) \big] = \int p_B(x) f(x) dx$
\end{itemize}

%\clearpage

%%%%%%%%%%%%%%%%%%%%%%%%%%%%%%%%%%%%%%%%%%%%%%%%%%%%%%%%%%%%%%%%%%%%%%%%%%%%%%%%%%%%%%%%%%%%%%%%%%%
%%%%%%%%%%%%%%%%%%%%%%%%%%%%%%%%%%%%%%%%%%%%%%%%%%%%%%%%%%%%%%%%%%%%%%%%%%%%%%%%%%%%%%%%%%%%%%%%%%%
\section{Derivation of \texorpdfstring{$\KL(p_G||p_B)$}{KL(pG||pB)}}
\label{sec:appendix:KLx}

{%\small
    \begin{subequations}
        \label{eq:KLx:KLx_dvt}
        \begin{align}
            \KL(p_G || p_B) & = \int p_G(x) \log \frac{p_G(x)}{p_B(x)} \ dx \label{subeq:KLx:KLx_dvt_a} \\
            & = \int p_G(x) \log\cZ_B \ dx + \int p_G(x) \log \frac{p_G(x)}{\tilde{p}_B(x)} \ dx \label{subeq:KLx:KLx_dvt_b} \\
            & = \log\cZ_B + \int p_G(x) \log \frac{p_G(x)}{\tilde{p}_B(x)} \ dx \label{subeq:KLx:KLx_dvt_c} \\
            & = \log\cZ_B + \int q_\cN(z) \log \frac{q_\cN(z) \cdot \left| \det \left( \frac{\partial G(z)}{\partial z} \right) \right|^{-1}}{\tilde{p}_B(G(z))} \ dz \label{subeq:KLx:KLx_dvt_d} \\
            & = \log\cZ_B - S_\cN + \int q_\cN(z) \log \frac{\left| \det \left( \frac{\partial G(z)}{\partial z} \right) \right|^{-1}}{\tilde{p}_B(G(z))} \ dz \label{subeq:KLx:KLx_dvt_e} \\
            & = \log\cZ_B - S_\cN + \int q_\cN(z) \log \frac{\left| \det \left( \frac{\partial G(z)}{\partial z} \right) \right|^{-1}}{e^{-\beta U_B(G(z))}} \ dz \label{subeq:KLx:KLx_dvt_f} \\
            & = \log\cZ_B - S_\cN + \underset{z_\cN \sim q_\cN}{\bbE} \left[ \beta U_B(G(z_\cN)) + \log \left| \det \left( \frac{\partial G(z_\cN)}{\partial z_\cN} \right) \right|^{-1} \right] \label{subeq:KLx:KLx_dvt_g} \\
            & = \log\cZ_B - S_\cN + \underset{z_\cN \sim q_\cN}{\bbE} \left[ \beta U_B(G(z_\cN)) - \log \left| \det \left( \frac{\partial G(z_\cN)}{\partial z_\cN} \right) \right| \right] \label{subeq:KLx:KLx_dvt_h}
        \end{align}
    \end{subequations}
}

with:
\begin{itemize}[noitemsep, topsep=0mm]
    \item[-] (\ref{subeq:KLx:KLx_dvt_a}) by definition of the $\KL$ divergence
    \item[-] (\ref{subeq:KLx:KLx_dvt_b}) by using $p_B = \tilde{p}_B / \cZ_B$
    \item[-] (\ref{subeq:KLx:KLx_dvt_c}) by using $\int p_G(x) dx = 1$ (probabilities sum to one)
    \item[-] (\ref{subeq:KLx:KLx_dvt_d}) by using the change of variable formula:
    \begin{equation}
        \label{eq:UG:change_of_var}
        \begin{aligned}
            p_G(x) \ dx &= q_\cN(G^{-1}(x)) \cdot \left| \det \left( \frac{\partial G^{-1}(x)}{\partial x} \right) \right| \ dx \\
            &= q_\cN(z) \cdot \left| \det \left( \frac{\partial G(z)}{\partial z} \right) \right|^{-1} \ dx \\
            &= q_\cN(z) \ dz
        \end{aligned}
    \end{equation}
    \item[-] (\ref{subeq:KLx:KLx_dvt_e}) by definition of the entropy: $S_\cN = S(q_\cN) = - \int q_\cN(x) \log q_\cN(x) dx$
    \item[-] (\ref{subeq:KLx:KLx_dvt_f}) by using: $\tilde{p}_B(x) = e^{-\beta U_B(x)}$
    \item[-] (\ref{subeq:KLx:KLx_dvt_g}) by definition of expectation: $\underset{z_\cN \sim q_\cN}{\bbE} \big[ f(z_\cN) \big] = \int q_\cN(z) f(z) dz$
\end{itemize}

%\clearpage

%%%%%%%%%%%%%%%%%%%%%%%%%%%%%%%%%%%%%%%%%%%%%%%%%%%%%%%%%%%%%%%%%%%%%%%%%%%%%%%%%%%%%%%%%%%%%%%%%%%
%%%%%%%%%%%%%%%%%%%%%%%%%%%%%%%%%%%%%%%%%%%%%%%%%%%%%%%%%%%%%%%%%%%%%%%%%%%%%%%%%%%%%%%%%%%%%%%%%%%
\section{Optimizing \texorpdfstring{$\cL_{\KL x}$}{KLx} leads to mode collapse on \texttt{Dialanine}}
\label{sec:appendix:KLx_dialanine_collapse}

\begin{figure}[!ht]
    \centering
    \begin{subfigure}{1\textwidth}\hspace{.35cm}
        \includegraphics[width=380pt, height=68pt]{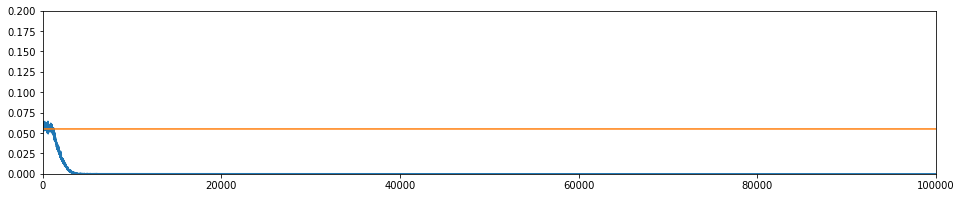}
        \caption{Percentage of generated samples $x_G \sim p_G$ in the minor mode during fine-tuning with $\cL_{\KL x}$.}
        \label{fig:PP_dialanine_lambda10_KLx_finetuning_ratios}
    \end{subfigure}
    \begin{subfigure}{.32\textwidth}\centering
        \includegraphics[width=120pt, height=68pt]{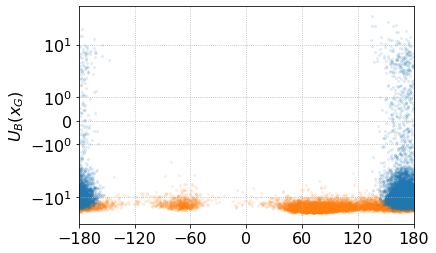}
        \caption{$U_B$ energies}
        \label{fig:PP_dialanine_lambda10_KLx_UBxG}
    \end{subfigure}
    \begin{subfigure}{.32\textwidth}\centering
        \includegraphics[width=90pt, height=68pt]{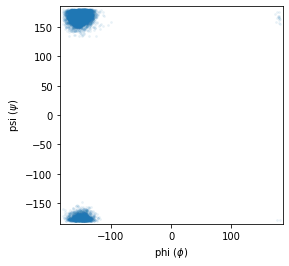}
        \caption{$p_G$ projection on $(\phi, \psi$)}
        \label{fig:PP_dialanine_lambda10_KLx_ramachandran}
    \end{subfigure}
    \begin{subfigure}{.32\textwidth}\centering
        \includegraphics[width=120pt, height=68pt]{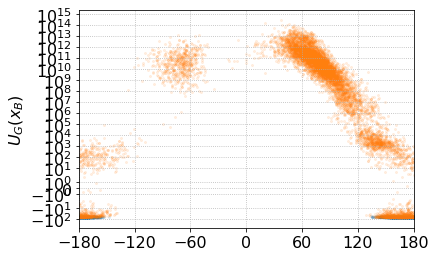}
        \caption{$U_G$ energies}
        \label{fig:PP_dialanine_lambda10_KLx_UGxB}
    \end{subfigure}
    \caption{Results of the fine-tuning with $\cL_{\KL x}$ on \texttt{Dialanine}. See the captions of figure 3 of the main paper for more details.}
    \label{fig:KLs:double_well_12Dmedium_pretraining}
\end{figure}

Although most generated samples have low energy, not all of them do (figure~\ref{fig:PP_dialanine_lambda10_KLx_UBxG}) and they only represent a subset of the target distribution (figure~\ref{fig:PP_dialanine_lambda10_KLx_UGxB}), since during training minor modes are progressively lost (figure~\ref{fig:PP_dialanine_lambda10_KLx_finetuning_ratios}), until a single mode remains in the 2D projection (figure~\ref{fig:PP_dialanine_lambda10_KLx_ramachandran}).

%\clearpage

%%%%%%%%%%%%%%%%%%%%%%%%%%%%%%%%%%%%%%%%%%%%%%%%%%%%%%%%%%%%%%%%%%%%%%%%%%%%%%%%%%%%%%%%%%%%%%%%%%%
%%%%%%%%%%%%%%%%%%%%%%%%%%%%%%%%%%%%%%%%%%%%%%%%%%%%%%%%%%%%%%%%%%%%%%%%%%%%%%%%%%%%%%%%%%%%%%%%%%%
\section{Derivation of \texorpdfstring{$\KL(q_F||q_\cN)$}{KL(qF||qN)} with Importance Sampling}
\label{sec:appendix:KLz_bws}

{%\small
    \begin{subequations}
        \label{eq:KLz_bws:KLz_xGd_dvt}
        \begin{align}
            \nabla_\theta \KL & (q_F||q_\cN) = \nabla_\theta \left[ \underset{x_B \sim p_B}{\bbE} \left[ \frac{1}{2\sigma^2} U_\cN(F(x_B)) - \log \left| \det \left( \frac{\partial F(x_B)}{\partial x_B} \right) \right| \right] \right] \label{subeq:KLz_bws:KLz_xGd_dvt_a} \\
            & = \nabla_\theta \left[ \int p_B^\ddagger(x) \left( \frac{1}{2\sigma^2} U_\cN(F(x)) - \log \left| \det \left( \frac{\partial F(x)}{\partial x} \right) \right| \right) dx \right] \label{subeq:KLz_bws:KLz_xGd_dvt_b} \\
            & = \nabla_\theta \left[ \int p_G^\ddagger(x) \left( \frac{p_B(x)}{p_G(x)} \right)^\ddagger \left( \frac{1}{2\sigma^2} U_\cN(F(x)) - \log \left| \det \left( \frac{\partial F(x)}{\partial x} \right) \right| \right) dx \right] \label{subeq:KLz_bws:KLz_xGd_dvt_c} \\
            & = \nabla_\theta \left[ \underset{x_G^\ddagger \sim p_G^\ddagger}{\bbE} \left[ \left( \frac{p_B(x_G^\ddagger)}{p_G(x_G^\ddagger)} \right)^\ddagger \left( \frac{1}{2\sigma^2} U_\cN(F(x_G^\ddagger)) - \log \left| \det \left( \frac{\partial F(x_G^\ddagger)}{\partial x_G^\ddagger} \right) \right| \right) \right] \right] \label{subeq:KLz_bws:KLz_xGd_dvt_d} \\
            & = \frac{1}{\cZ_B} \cdot \nabla_\theta \left[ \underset{x_G^\ddagger \sim p_G^\ddagger}{\bbE} \left[ \left( \frac{\tilde{p}_B(x_G^\ddagger)}{p_G(x_G^\ddagger)} \right)^\ddagger \left( \frac{1}{2\sigma^2} U_\cN(F(x_G^\ddagger)) - \log \left| \det \left( \frac{\partial F(x_G^\ddagger)}{\partial x_G^\ddagger} \right) \right| \right) \right] \right] \label{subeq:KLz_bws:KLz_xGd_dvt_e} \\
            & = \frac{1}{\cZ_B} \cdot \underset{x_G^\ddagger \sim p_G^\ddagger}{\bbE} \nabla_\theta \left[ \left( \frac{\tilde{p}_B(x_G^\ddagger)}{p_G(x_G^\ddagger)} \right)^\ddagger \left( \frac{1}{2\sigma^2} U_\cN(F(x_G^\ddagger)) - \log \left| \det \left( \frac{\partial F(x_G^\ddagger)}{\partial x_G^\ddagger} \right) \right| \right) \right] \label{subeq:KLz_bws:KLz_xGd_dvt_f}
        \end{align}
    \end{subequations}
}

with:
\begin{itemize}[noitemsep, topsep=0mm]
    \item[-] (\ref{subeq:KLz_bws:KLz_xGd_dvt_a}) by taking the gradient of equation~\ref{subeq:KLz:dvt_h}.
    \item[-] (\ref{subeq:KLz_bws:KLz_xGd_dvt_b}) by definition of expectation: $\underset{x_B \sim p_B}{\bbE} \big[ f(x_B) \big] = \int p_B(x) f(x) dx$. Note the subtle replacement of $p_B$ with $p_B^\ddagger$ which is allowed inside the gradient operator $\nabla_\theta$ since $p_B$ is not a function of $\theta$.
    \item[-] (\ref{subeq:KLz_bws:KLz_xGd_dvt_c}) by \href{https://en.wikipedia.org/wiki/Importance_sampling}{importance sampling}.
    \item[-] (\ref{subeq:KLz_bws:KLz_xGd_dvt_d}) by definition of expectation.
    \item[-] (\ref{subeq:KLz_bws:KLz_xGd_dvt_e}) by definition of $\tilde{p}_B = \cZ_B p_B$.
    \item[-] (\ref{subeq:KLz_bws:KLz_xGd_dvt_f}) by noticing that, although $p_G^\ddagger$ is a function of $\theta$, it is not a \textit{differentiated} function of $\theta$. Since it is detached, it is treated as a constant by the gradient operator and the expectation can be sampled in the context of Stochastic Gradient Descent.
\end{itemize}
%\clearpage

%%%%%%%%%%%%%%%%%%%%%%%%%%%%%%%%%%%%%%%%%%%%%%%%%%%%%%%%%%%%%%%%%%%%%%%%%%%%%%%%%%%%%%%%%%%%%%%%%%%
%%%%%%%%%%%%%%%%%%%%%%%%%%%%%%%%%%%%%%%%%%%%%%%%%%%%%%%%%%%%%%%%%%%%%%%%%%%%%%%%%%%%%%%%%%%%%%%%%%%
\section{Analysis of the bias when generating deterministic, minimum-energy hydrogen coordinates}
\label{sec:appendix:mode_ratio_change}

In our two-stage architecture, the normalizing flow generator $G$ outputs only heavy atom coordinates $x^C$, while hydrogen atoms are added at minimum-energy positions $x^H$ by an auxiliary neural network denoted by $h$.
Thus, all-atom coordinates are generated as $\{x^C, x^H\} = h(x^C) = h(G(z))$, and the reverse operation is $z = F(\bar{h}(x^C, x^H))$, noting $\bar{h}$ the operation of stripping H coordinates from an all-atom configuration.

As a result, whereas the generator $G$ is bijective, the complete pipeline $h\circ G$ is not:
while $F \circ \bar{h}\circ h\circ G$ is identity in the latent space, $h\circ G \circ F \circ \bar{h} = h\circ\bar{h}$ corresponds to energy minimization with respect to hydrogen atom coordinates, i.e. the \textit{projection} of complete atomic coordinates onto the minimum-energy-hydrogen manifold.

In the spirit of a coarse-graining (CG) approach, the desirable target for the generated distribution $p_G^C$ of heavy atom coordinates is the marginal $p_B^C$ of the target $p_B$ with respect to those coordinates:
\begin{equation}
    p_B^C(x^C) = \int p_B(x^C, x^H) dx^H
\end{equation}

We characterize convergence on the dialanine example by computing the predicted probability of the minor mode  $\cM$ of dialanine (known to biochemists as the C7ax conformation).
The Boltzmann probability of this mode is:
\begin{align}
P_B(\cM) &= \int_\cM p_B(x) dx \\
         &= \int_\cM p_B(x^C, x^H) dx^C dx^H
\end{align}
Now we use the fact that $\cM$ is defined solely based on the values of $x^C$, so that it can be written $\cM = \cM^C \times \R^{3N_H}$, with $\cM^C$ a set of heavy atom coordinates, and $N_H$ the number of hydrogen atoms.
\begin{align}
P_B(\cM) &= \int_{\cM^C} \left[ \int p_B(x^C, x^H) dx^H \right]dx^C \\
                 &= \int_{\cM^C} p_B^C(x^C) dx^C
\end{align}

However, in practice, the optimization of $G$ minimizes the divergence between $p_G^C$ and an auxiliary distribution $\bar{p}_B$ defined by:
\begin{equation}
    \bar{p}_B(x^C) = p_B(h(x^C))
\end{equation}
Thus any difference between $p_B^C$ and $\bar{p}_B$ introduces a bias in the generation of heavy atom coordinates.
Furthermore, the probability $p_G$ of generation of an all-atom configuration $x$ is:
\begin{equation}
    p_G(x) = p_G^C(\bar{h}(x)) \; \delta(x-h\circ\bar{h}(x))
\end{equation}
which is non-zero only on the minimum-energy-hydrogen manifold that is the image of $h\circ\bar{h}$.

Assuming perfect training ($p_G^C = \bar{p_B}$), we obtain:
\begin{equation}
    p_G(x) = p_B(h\circ\bar{h}(x)) \delta(x-h\circ\bar{h}(x))
\end{equation}

The probability of the minor mode as generated by a perfectly trained network is thus:
\begin{align}
\bar{P}_B(\cM) &= \int_{\cM} p_G(x) dx \\
             &= \int_\cM  p_B(h\circ\bar{h}(x)) \delta(x-h\circ\bar{h}(x)) dx \\
            &\approx \int_\cM  p_B(h\circ\bar{h}(x)) dx
\end{align}
where the last step relies on the fact that the conditional Boltzmann distribution of hydrogen atom positions is peaked, that is, $p_B$ is largest around minimal-energy hydrogen coordinates (where $x = h\circ\bar{h}(x)$).

This leads to an importance sampling estimator for this probability based on samples from the reference dataset:
\begin{align}
    \label{eq:crits:dialanine_expected_minor_mode_ratio}
    \hat{\bar{P}}_B(\cM):= \frac{\displaystyle \sum_{x_B \sim p_B \vert x_B \in \cM} \frac{\tilde{p}_B(h\circ\bar{h}(x_B))}{\tilde{p}_B(x_B)}}{\displaystyle \sum_{x_B \sim p_B} \frac{\tilde{p}_B(h\circ\bar{h}(x_B))}{\tilde{p}_B(x_B)}} \approx 1.21\%
\end{align}

which we use as a reference value in Figure~\ref{fig:PP_dialanine_lambda10_finetuning_ratios}.

%\clearpage

\section{Details and proofs for section 3.1 (Estimator variance as a loss)}
\label{appendix:proof_variance_loss}

\subsection{Integral quantity of interest}

Let us suppose that the use case of our generator is to estimate integral quantities of the form:
$$ \E_{p_B}[ f ]$$
for some given function(s) $f$.

Let us denote by $Q$ its true value.
$$ Q := \E_{p_B}[ f ] := \int_{x \in X} f(x)\; p_B(x) \,dx \;=\; \E_{p_G}\left[ f \frac{p_B}{p_G} \right]$$
This is exact provided that $p_G$ is never 0 where $p_B$ is not.

\subsection{Estimation by sampling}

In practice, one estimates $Q$ by sampling:
%$\newcommand\hQ{\hat{Q}}$
$$ Q \simeq \hQ := \frac{1}{N} \sum_{x_i \in m} f(x_i) \frac{p_B(x_i)}{p_G(x_i)}$$
where $m = (x_1, \dots, x_N)$ is a mini-batch of points sampled according to $p_G$.
Note however than we do not know $p_B$, but only $\tp_B = \cZ_B p_B$. We will come back to this point later.

\subsection{This estimator is unbiased}

Whatever $p_G$, $\hQ$ is an approximation of $Q$, in that for very large mini-batches $m$, i.e. large $N$, the estimate $\hQ$ tends to $Q$. One then says that the estimator is \emph{unbiased}. The convergence rate is typically in $O(1/\sqrt{N})$.
Indeed:
$$\E_{m \,\sim\, p_G^N} \;[\; \hQ \;] = Q$$
where the expectation is taken over mini-batches of $N$ independent samples, taken according to $p_G$.
To prove this, see that even for just one sample ($N = 1$) one has:
$$ \hQ = f(x_1) \frac{p_B(x_1)}{p_G(x_1)} $$
and thus:
$$\E_{x_1 \,\sim\, p_G} \; \left[\, \hQ \;\right] \;=\; \E_{x \,\sim\, p_G} \; \left[\, f(x) \frac{p_B(x)}{p_G(x)} \,\right] \;=\; \int_{x \in X} f(x)\, p_B(x)\,dx \;=:\; Q$$
For a mini-batch of arbitrary size $N$, one gets the average of $N$ such quantities, each of which are $Q$ on expectation, so one recovers $Q$ again:
\begin{align*}
    \E_{m  \sim  p_G^N}  [  \hQ  ] &= \E_{m  \sim  p_G^N}  \left[  \frac{1}{N} \sum_{x_i \in m} f(x_i) \frac{p_B(x_i)}{p_G(x_i)}   \right] \\
    &= \frac{1}{N} \sum_{i = 1}^N \E_{x_i  \sim  p_G}  \left[   f(x_i) \frac{p_B(x_i)}{p_G(x_i)}   \right] \\
    &=  \frac{1}{N} \sum_{i = 1}^N \E_{x  \sim  p_G}  \left[   f(x) \frac{p_B(x)}{p_G(x)}   \right]
\end{align*}

because points $x_i$ are sampled independently;
$$=  \E_{x \,\sim\, p_G} \;\left[\;  f(x) \frac{p_B(x)}{p_G(x)}  \;\right] \; =: \; Q$$

\subsection{Variance of the estimator}

Yet, for some distributions $p_G$, the estimate $\hQ$ may converge faster than for other ones, in terms of number of samples required to reach a given target accuracy. This is reflected in the \emph{variance} of the estimator $\hQ$:

$$V = \E_{m \,\sim\, p_G^N}\;[\; (\hQ - Q)^2 \;]$$

that one would like to be as small as possible. Indeed the typical gap between an estimate $\hQ$ for a mini-batch and the real value $Q$ can be expected to be of the order of magnitude of $V$ (by definition).

Can we train $p_G$ so as to minimize $V$?

\subsection{Reducing variances over mini-batches to variances over single samples}

For a given mini-batch size $N$, the variance over the choice of mini-batch $m$ is:

\begin{align*}
    V &=  \E_{m  \sim  p_G^N} [  (\hQ - Q)^2  ]\\
    &= \E_{m  \sim  p_G^N} [  \hQ^2  ] - Q^2
\end{align*}

As $Q^2$ is constant (does not depend on $p_G$), we aim at minimizing only:
\begin{align*}
    \E_{m  \sim  p_G^N} [  \hQ^2  ] &=  \E_{m  \sim  p_G^N} \left[  \left( \frac{1}{N} \sum_{x_i \in m} f(x_i) \frac{p_B(x_i)}{p_G(x_i)} \right)^2  \right] \\
    &=   \frac{1}{N^2} \E_{m  \sim  p_G^N} \left[   \sum_{x_i \in m} f^2(x_i) \frac{p_B^2(x_i)}{p_G^2(x_i)}   + \sum_{x_i, x_j \in m i\neq j} f(x_i) \frac{p_B(x_i)}{p_G(x_i)} f(x_j) \frac{p_B(x_j)}{p_G(x_j)}   \right]
\end{align*}

Note that points $x_i$ and $x_j$ are sampled independently, and all points are sampled identically (according to the same law), and thus:
\begin{align*}
    \E_{m  \sim  p_G^N} [ \hQ^2 ] & = \frac{1}{N^2} \sum_{i=1}^N \E_{x_i  \sim  p_G} \left[ f^2(x_i) \frac{p_B^2(x_i)}{p_G^2(x_i)} \right]  +  \frac{1}{N^2} \sum_{i,j=1, i\neq j}^N \E_{x_i  \sim  p_G} \left[  f(x_i) \frac{p_B(x_i)}{p_G(x_i)} \right] \E_{x_j  \sim  p_G} \left[  f(x_j) \frac{p_B(x_j)}{p_G(x_j)}  \right] \\
    & =  \frac{1}{N} \E_{x  \sim  p_G} \left[ f^2(x) \frac{p_B^2(x)}{p_G^2(x)} \right]  +  \frac{N(N-1)}{N^2} \E_{x \sim  p_G} \left[ f(x) \frac{p_B(x)}{p_G(x)} \right]^2 \\
    & =  \frac{1}{N} \E_{x  \sim  p_G} \left[ f^2(x) \frac{p_B^2(x)}{p_G^2(x)} \right]  +  \left(1 - \frac{1}{N}\right)  Q^2
\end{align*}

The variance (without forgetting any constant term) thus interestingly rewrites as:
$$V = \frac{1}{N} \left( \E_{x  \sim  p_G} \left[  f^2(x) \frac{p_B^2(x)}{p_G^2(x)}  \right]   -   Q^2 \right) $$
which can be interpreted as: the variance of an estimator based on $N$ samples is $\frac{1}{N}$ times the variance of the estimator based on a single sample. This implies that the variance behaves as $O(\frac{1}{N})$ and thus the typical error (standard deviation) is $O(\frac{1}{\sqrt{N}})$.

\subsection{Optimizing the variance w.r.t. \texorpdfstring{$p_G$}{pG}}

Based on the variance formula above, one can consider that the quality (or rather: expected error) of the generator $G$ can be quantified as:
$$C(p_G) \;=\; \E_{x \,\sim\, p_G}\;\left[\; f^2(x) \frac{p_B^2(x)}{p_G^2(x)} \;\right] $$
and we would like to minimize it w.r.t. $p_G$.

If $f$ can be any bounded function over the space $X$ of points $x$, then one can deduce the following optimization criterion:
$$C(p_G) \;=\; \E_{x \,\sim\, p_G}\;\left[\; \frac{p_B^2(x)}{p_G^2(x)} \;\right] $$

Note that this resembles a $\KL$ divergence without the logarithm, and is also equal to:
$$C(p_G)  =  \E_{x  \sim  p_G} \left[  e^{2 \beta  (U_G - U_B)} \right] \frac{1}{\mathcal{Z}_B^2}$$

This loss is also equal to:
$$C(p_G) \;=\; \E_{x \,\sim\, p_B}\;\left[\; \frac{p_B(x)}{p_G(x)} \;\right] $$
though this is not directly exploitable.

Note that if function $f$ that needs to be integrated is known, it should be used explicitly in the criterion to optimize!

\subsection{Special case: estimating free energy differences}

An interesting and classic case in practice is to compute the free energy difference $\Delta F_{BC}$ between the sate being sampled (with energy $U_B$) and an alternate state defined by energy $U_C$. Then:
$$f(x) = e^{-\beta(U_C(x)-U_B(x))} $$

and $$ e^{-\beta \Delta F_{BC}} = \E_{x  \sim  p_B} [f] $$

%\clearpage

%%%%%%%%%%%%%%%%%%%%%%%%%%%%%%%%%%%%%%%%%%%%%%%%%%%%%%%%%%%%%%%%%%%%%%%%%%%%%%%%%%%%%%%%%%%%%%%%%%%
%%%%%%%%%%%%%%%%%%%%%%%%%%%%%%%%%%%%%%%%%%%%%%%%%%%%%%%%%%%%%%%%%%%%%%%%%%%%%%%%%%%%%%%%%%%%%%%%%%%
\section{Proof of L2 loss pseudo-gradient descent convergence (section 4)} %\ref{sec:PL2})}
\label{sec:appendix:L2_loss_convergence}

We study here the optimization properties of the masked $L^2$ loss, with detached means.

\subsection{Notations}

Given a dataset of points $x_i$, we denote by
$$r_i = \log \frac{p_B(x_i)}{p_G(x_i)}$$
the log-ratio of the target and generated densities at point $x_i$.

Differences of log-ratios satisfy:
$$r_i - r_j = \log \frac{p_B(x_i)}{p_G(x_i)} - \log \frac{p_B(x_i)}{p_G(x_i)}
= \log \frac{\tp_B(x_i)}{p_G(x_i)} - \log \frac{\tp_B(x_i)}{p_G(x_i)}$$
where $\tp_B = (\log \cZ_B) \,  p_B$ is easily computable, which makes such differences easily computable, to the opposite of the log-ratios $r_i$ themselves.

Let us note that:
$$\E_{x_j\sim p_B}[r_j] = \KL(p_B||p_G)$$
and similarly:
$$\E_{x_i\sim p_G}[r_i] = - \KL(p_G||p_B)$$

\subsection{Pairwise L2 loss}

\subsubsection{Definition}

The pairwise $L^2$ loss (simple version) is defined as:
$$L(p_G,p_B) = \var_{x_i\sim p_G}(r_i) = \E_{x_i\sim p_G}\left[ \left(r_i - \E_{x_j\sim p_G}[r_j] \right)^2 \right]$$

As a side remark, let us note that this is equal to:
$$L(p_G,p_B) = \frac{1}{2}\E_{x_i, x_j \sim p_G}\left[ (r_i - r_j)^2 \right]$$
but we will not use this property here.
The proofs are the same as for the mixed sampling case (cf below).

\subsubsection{Global minimum}

This loss is also equal to:
$$L(p_G,p_B) = e^{D_2(p_G||p_B)}$$
where $D_2$ is the Rényi divergence of order 2, a measure of divergence between distributions.
$D_2$ is a $f$-divergence and in particular it is jointly convex.
As a consequence, the only minimum is the global one, reached at $p_G=p_B$.

\subsection{Masked L2 loss with detached means}

\subsubsection{Definition}

The masked $L^2$ loss (simple version) with detached means is defined as:
$$L_{MD}(p_G,p_B) = \E_{x_i\sim p_G}\left[ \left(r_i -  K^\ddagger \right)_+^2\right]$$
where $a_+^2 = a^2$ if $a>0$ and 0 otherwise,
and where $K = \E_{x_j\sim p_G}[r_j] = -\KL(p_G||p_B) \leqslant 0$ is not differentiated (considered as a constant at every time step of the gradient descent); we say $K^\ddagger$ is \textit{detached}, following PyTorch vocabulary.

This definition is motivated as follows:
\begin{itemize}
\item masking $(r_i - K)$ with $()_+$ to consider it only when positive has the consequence that $r_i$ will be only asked to decrease.  Since $p_G$ is a probability distribution, this implies that some other $r_j$ will increase to compensate ($p_G$ cannot increase everywhere), but at least this will not be done by the gradient descent, hence not in the worst possible direction (make $x_j$ as unlikely as possible, and this as fast as possible, i.e. make it as unrealistic as possible), but rather in the smoothest possible way (push the probability mass to regions where it is more needed).
\item not detaching $K$ would ask it to increase, and thus to decrease values of $p_G$ at most points $x_j$.
\end{itemize}

\subsubsection{Global minimum}

This loss is non-negative, and 0 can be reached if all $r_i$ are equal (note that 0-loss implies that no $r_i$ is greater than the mean $K$, and consequently no $r_i$ can be strictly less than the mean $K$ as well, otherwise the mean would be lower than itself).
As previously, this is the case if and only if $p_G$ is proportional to $\tp_B$ and thus equal to $p_B$ (see the end of the convergence proof below for more details).

\subsubsection{Optimization by partial gradient descent}

However, since only part of the loss is differentiated ($K$ is considered a constant though changing with $p_G$), then strictly speaking the optimization process is not a gradient descent, as the loss is different at each time step. Therefore one needs to check that this optimization process does converge, and to the global minimum.

This task is hindered by the fact that the constraint that the total mass of $p_G$ has to remain 1 is handled implicitly by the normalizing flow, and so the precise way the gradient w.r.t. $p_G$ is replaced with a variation $\delta p_G$ that preserves the total mass depends on the architecture and the neural network weights.

\paragraph{Total variation of log-ratios}

Let us study the variation of $K$ induced by a (partial) gradient step:
$$\delta K = \delta \left(\E_{x\sim p_G}[r]\right)  = \delta \left( \int p_G\, r\, dx \right) = \E_{x\sim p_G}[\delta r] + \int r \,\delta p_G\, dx$$

Note that:
$$p_G(x_i) = e^{\log p_G(x_i)} = e^{-r_i + \log p_B(x_i)} = e^{-r_i} p_B(x_i)$$

As $\E_{x\sim p_G}[1] = 1$ we have $\E_{x\sim p_B}[e^{-r(x)}] = 1$, and this for any $p_G$ or equivalently for any associated $r$.
As a consequence, the variation of $\E_{x\sim p_B}[e^{-r}]$ w.r.t. to any realizable change $\delta r$ ($p_B$ being fixed and $p_G$ varying) is necessarily 0:
$$\delta\left(\E_{x\sim p_B}[e^{-r}]\right) = \E_{x\sim p_B}[e^{-r}\delta r] = 0$$
which rewrites as
$$\E_{x\sim p_G}[\delta r] = 0$$

Consequently $\delta K$ can be simplified as:
$$\delta K =  \int r \,\delta p_G\, dx$$
and rewritten as:
$$\delta K =  \int (r-K) \,\delta p_G\, dx$$
since $K \int \delta p_G = 0$ as $p_G$ has conserved total mass = 1.

Now, note that the gradient descent step is asking to decrease all $r$ that are greater than $K$. Since $r = \log p_B/p_G$, this means increasing $p_G$ for such points where $r>K$.
So $\delta p_G > 0$ when $r-K>0$.

For points where $r<K$, $p_G$ is not asked to change, but the conservation of mass makes that (at least some of) such points will have their probabilities decreased, to produce the extra mass needed by the previous points above ($r>K$). Thus $\delta p_G \leqslant 0$ where $r<K$.

In the end, for all points, $(r-K) \delta p_G \geqslant 0$  and consequently:
$$\delta K \geqslant 0$$

See Section \ref{sec:optimmoredetails} below for more details about this proof.

\paragraph{Consequences}

Therefore, $K$ is increasing with time.
This is good news, as $K = -\KL(p_G||p_B)$: this means that with this optimization process, $p_G$ is getting closer to $p_B$.

As $K$ is actually strictly increasing as long as $p_G$ is not $p_B$, we can conclude that the optimization process will lead to the desired global optimum.

Another way to see this is that $K$ is an increasing, upper-bounded value (bounded by 0) and this will converge. When $K$ converges (possibly to a non-0 value), then the training criterion becomes stable with time and the optimization process becomes a real gradient descent w.r.t. $r_i$.
Therefore a local minimum of this loss (for fixed limit $K$) will be reached. Now, the gradient of this fixed-K loss is 0 when all $r_i$ are either equal to or less than $K$. If at such a minimum one $r_i$ was strictly less than $K$, we would get that the average of all $r_i$ (according to $p_G$) would be strictly less than $K$, while it is precisely $K$. Therefore all $r$ are equal and the global minimum is reached.

\subsubsection{More details on the proof}
\label{sec:optimmoredetails}

We explain here in more details the links between the signs of $\delta p_G(x)$ and of $(r(x)-K)$, that is, that their product is never negative.

To see this, we need to detail how $\delta p_G(x)$ is obtained. We are optimizing the following criterion :

$$L_{MD}(p_G,p_B) = \E_{x_i\sim p_G}\left[ \left(r_i -  K^\ddagger \right)_+^2\right]$$

with $K = \E_{x_j\sim p_G}[r_j] = -\KL(p_G||p_B)$,
where:
\begin{itemize}
\item the sampling distribution $p_G$ over which the expectation is performed is not differentiated, i.e. the minibatch points $x_G = (x_i)$ are detached; this is similar to what is done with VarGrad~\cite{Richter2020} ;
\item the log ratios $r_i = \log \frac{p_B(x_i)}{p_G(x_i)}$ are differentiated, with respect to the parameters of the modeled distribution $p_G$, and this will induce a desired variation for $p_G$ ;
\item $K$ is not differentiated ;
\item only points for which $r_i > K$ are actually taken into account in the criterion.
\end{itemize}

These two last points differ from VarGrad~\cite{Richter2020}; practice shows that they are required for the optimization process to go well.
Interestingly, this pseudo gradient descent can be proven theoretically to converge towards the right minimum. We will prove that $K = -\KL(p_G||p_B)$ increases with time, and tends to 0, and thus $p_G$ gets closer to $p_B$ at each step and finally converges to the target.

The hypotheses for this theoretical study are only that:
\begin{itemize}
\item the support of $p_G$ includes the one of $p_B$ ;
\item the neural network is expressive enough (for the pseudo gradient direction to be followed).
\end{itemize}

\paragraph{Desired variations}

Let us first note that the log ratio is $r(x) = \log\frac{p_B(x)}{p_G(x)}$, i.e. $p_G = e^{-r} p_B$.
As a consequence, possible variations of $r$ or $p_G$ are linked as follows:
$$\delta r = - \frac{1}{p_G} \delta p_G$$
$$\delta p_G = - p_G \, \delta r$$

The (opposite of the) derivative of $L_{MD}(p_G,p_B)$ with respect to $r$ yields the desired variation:
$\delta r_\text{desired}(x) = - 2 (r(x) - K)_+ \, p_G(x)$
having taken into account that only some parts of the criterion are differentiated as explained above.
This translates into a desired distribution variation :
$\delta {p_G}_\text{desired} = 2 (r - K)_+ \, p_G^2$
This is 0 for points $x$ such that $r(x) \leqslant K$ and positive otherwise.

\paragraph{Constrained variations}

However $p_G$ is a probability distribution and is constrained to sum up to 1. 
How a practical training step projects the desired probability variation $\delta {p_G}_\text{desired}$ onto a realizable variation $\delta {p_G}_\text{realizable}$ depends on the normalizing flow architecture, its weights, and its expressivity, in a complex manner. If the neural network is expressive enough though, the desired variation  $\delta {p_G}_\text{desired}$ is realizable up to the mass constraint. We will study here this ideal case, where the network is sufficiently expressive, and reason in the functional space of possible functions $p_G$, forgetting about the network (that will be able to express the realizable variation $\delta {p_G}_\text{realizable}$ anyway).

There are several ways to project $\delta {p_G}_\text{desired}$ onto a realizable variation $\delta {p_G}_\text{realizable}$ that satisfies the mass constraint. We do as follows:

For points $x$ such that  $r(x) > K$: 
$$\delta {p_G}_\text{realizable}(x)  = \delta {p_G}_\text{desired}(x)$$
and for other points:
$$\delta {p_G}_\text{realizable}(x)  = -\mu$$
where $\mu$ is the following constant (i.e. not depending on $x$):
$$\mu \;=\; \frac{1}{|\Omega^-|} \int_{x \in \Omega} \delta {p_G}_\text{desired}(x) \, dx \;\geqslant \;0$$
where $\Omega$ is the support of $p_G$ and where $\Omega^-$ is the subset of $\Omega$ where $r(x) < K$.
This construction is meant so that:
$$\int_{x \in \Omega} \delta {p_G}_\text{realizable}(x) \, dx \;=\; 0$$
which is the condition for $p_G$ to remain a probability distribution (the mass is kept constant).

Note that this projection of the desired pseudo-gradient over the set of variations satisfying the mass constraint is not the standard orthogonal one.

\paragraph{Impact on the average $K$}

By construction, the projected gradient will decrease the criterion $L_{MD}$ for fixed $K$ and fixed sampling distribution. However $p_G$ and consequently $K$ evolve with time. 

The variation of $K = \E_{x_j\sim p_G}[r_j]$ is:
$$\delta K = \delta\left( \int p_G\; r \right) = \int r \, \delta p_G + \int p_G \,\delta r$$

As explained earlier, the last term is 0: 
$\int p_G \, \delta r = \int - \delta p_G = 0$ for any realizable variation $\delta p_G$.

The variation of $K$ thus becomes:
$$\delta K = \int r \, \delta p_G = \int (r-K) \, \delta p_G = \int_{\Omega^+} (r-K) \delta p_G + \int_{\Omega^-} (r-K) \delta p_G$$
as $\int \delta p_G = 0$, and where $\Omega^+$ is the subset of $\Omega$ where $r(x) > K$.

Considering for $\delta p_G$ our pseudo gradient $\delta {p_G}_\text{realizable}$, we obtain:
$$\delta K = 2 \int_{\Omega^+} (r-K)^2 p_G^2 - \mu \int_{\Omega^-} (r-K)$$
where the fist term is non-negative, and $\mu \geqslant 0$ and $r-K<0$ on $\Omega^-$.
Consequently $\delta K \geqslant 0$.
Therefore $K$ keeps increasing with time.

Moreover,  $\delta K > 0$ as long as $p_G$ is not proportional to $p_B$ on the support of $p_G$.
As $K$ increases and is upper-bounded by $0$, $K$ converges. Convergence implies $\delta K = 0$, and therefore that $p_G$ is proportional to $p_B$ on the support of $p_G$. The hypothesis on the supports then implies that $p_G = p_B$.

%\clearpage

%%%%%%%%%%%%%%%%%%%%%%%%%%%%%%%%%%%%%%%%%%%%%%%%%%%%%%%%%%%%%%%%%%%%%%%%%%%%%%%%%%%%%%%%%%%%%%%%%%%
%%%%%%%%%%%%%%%%%%%%%%%%%%%%%%%%%%%%%%%%%%%%%%%%%%%%%%%%%%%%%%%%%%%%%%%%%%%%%%%%%%%%%%%%%%%%%%%%%%%
\section{Not detaching \texorpdfstring{$K$}{K}: experimental results}
\label{sec:appendix:detach_removal}

The masked $L^2$ loss with detached means is defined as:
\begin{equation}
    \label{eq:appendix:PL2_loss}
    \cL_{L^2_+}(\bx_G^\ddagger) = \sum_{i=1}^n \left[ \frac{1}{n} \cdot \left[ \left(r(x_{G, i}^\ddagger) -  K^\ddagger \right)_+^2\right] \right]
\end{equation}

\begin{figure}[!ht]
    \centering
    \begin{subfigure}{1\textwidth}\centering
        \includegraphics[width=300pt, height=80pt]{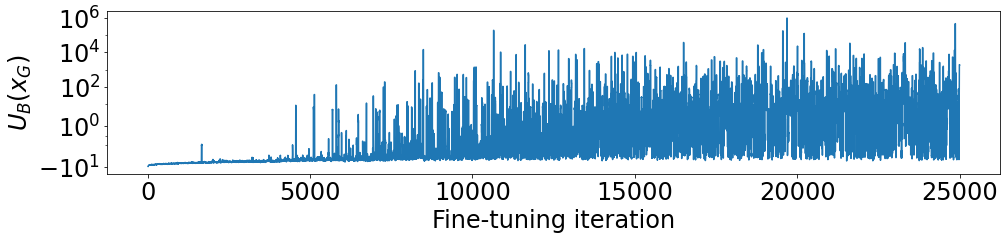}
        \caption{$U_B$ energy of generated samples $x_G \sim p_G$ during fine-tuning.}
        \label{fig:PP_dialanine_lambda10_PL2cachenodetach_finetuning_energies}
    \end{subfigure}
    \begin{subfigure}{.32\textwidth}\centering
        \includegraphics[width=120pt, height=80pt]{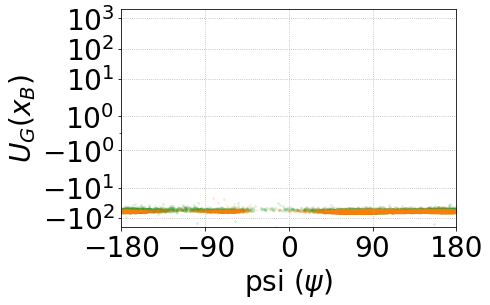}
        \caption{$U_B$ energies with $\cL_{L^2_+}$ with no detach of $K$}
        \label{fig:PP_dialanine_lambda10_PL2cachenodetach_UGxB}
    \end{subfigure}
    \begin{subfigure}{.32\textwidth}\centering
        \includegraphics[width=120pt, height=80pt]{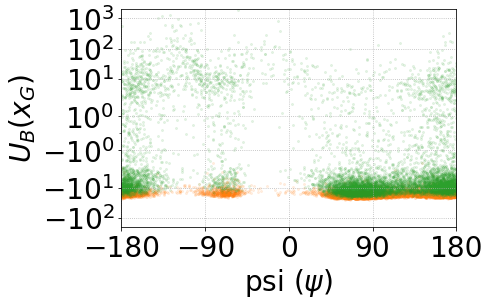}
        \caption{$U_G$ energies with $\cL_{L^2_+}$ with no detach of $K$}
        \label{fig:PP_dialanine_lambda10_PL2cachenodetach_UBxG}
    \end{subfigure}
    \caption{Results of a fine-tunings on \texttt{Dialanine} with a slightly modified version of $\cL_{L^2_+}$ where $K$ (from equation~\ref{sec:appendix:detach_removal}) is not detached. See caption of figure~\ref{fig:PP_dialanine_lambda10} of the main paper for more details.}
    \label{fig:KLs:PP_dialanine_lambda10_PL2cachenodetach}
\end{figure}

When $K$ from equation~\ref{eq:appendix:PL2_loss} is not detached, the potential energy $U_B$ of the generated samples $x_G$ is highly unstable during fine-tuning (\textcolor{PltBlue}{\ul{blue}} curve of figure~\ref{fig:PP_dialanine_lambda10_PL2cachenodetach_finetuning_energies} in this appendix), but when $K$ is detached the potential energies remain stable (\textcolor{PltGreen}{\ul{green}} curve of figure~\ref{fig:PP_dialanine_lambda10_finetuning_UBs} of the main paper). Very similar results are obtained when the mask of equation~\ref{eq:appendix:PL2_loss} is omitted, thereby illustrating experimentally that VarGrad~\cite{Richter2020} does not allow for stable data-free fine-tuning.

%\clearpage

%------------------------------

\section{Optimization pitfalls induced by the discretization of distributions into minibatches}
\label{sec:appendix:pitfalls}

In this section we list various optimization pitfalls encountered during gradient descents over a ``distances'' or divergences between probability distributions, and bring practical or theoretical recommendations to avoid them.

Historically for us, the results below were strong motivations to search for new optimization criteria with better optimization properties. We did not include this part in the main paper for space reasons and because these considerations are not essential, though they help understand the theoretical context of our study. The proofs and details are deferred to the next section, for readability reasons.

\subsection{Discretization issues with Kullback-Leibler and remedies}

To motivate this study of discretization and normalization issues, we start with an intruiguing fact.

\paragraph{Optimization of naively-discretized Kullback-Leibler does not converge towards the target.}
The main property of Kullback-Leibler divergence, as a measure of ``distance'' between probability distributions, is Gibbs inequality: $\KL(p_B||p_G) \geqslant 0$ for any $p_B, p_G$, with equality if and only if $p_B = p_G$.
Without the constraint of being probabilities (unit total mass), Gibbs inequality does not hold anymore, and thus minimizing $\KL(p_B||p_G)$ w.r.t. $p_G$ in the space of all distributions leads to an
unexpected behavior.

\begin{prop}[Unconstrained-mass pitfall] The gradient descent $\frac{dp_G}{dt} = - \nabla_{p_G} \KL(p_B||p_G)$, starting from $p_{G,0}$ and without constraining $\int_X p_G$ to be 1, yields, for large times $t$:
$$\forall x, \;\; p_{G,t}(x) \simeq \sqrt{2t} \sqrt{p_B(x)}$$
\end{prop}

Thus a lack of normalization will push $p_G$ to get infinite mass, and even correcting $p_{G,t}$ by its total mass will not yield $p_B$, but $\sqrt{p_B}$. Clearly, $p_G = p_B$ is not the minimizer of $\KL(p_B||p_G)$, and indeed with $p_{G,t} = \sqrt{2t}\sqrt{p_B}$ one gets $\KL(p_B||\sqrt{2tp_B}) = \frac{1}{2}H(p_B) - \frac{1}{2} \log (2t) \;\; <\!\!\!< \;0 = \KL(p_B||p_B)$ for high $t$.

This pitfall still stands even if one considers a parameterized model for $p_G$ that always satisfies the constraint $\int_X p_G = 1$, such as the output of a normalizing flow, if the discretization is inadequate.
Indeed the above also applies to continuous distributions discretized on a minibatch $m$ of samples:
$$p_{B|m}(x) := \sum_{i\in m} \delta_{x = x_i} p_B(x_i) \;\;\;\;\;\;\text{  and  }\;\;\;\;\;\; p_{G|m}(x) := \sum_{i\in m} \delta_{x = x_i} p_G(x_i)$$
The total mass of these distributions is not 1, even if normalized by the number of samples (minibatch size). It can be arbitrarily high, as $p_G(x_i)$ is a probability density.
As a consequence:
\begin{coro}[Unproper-minibatch-normalization pitfall] A gradient descent w.r.t. $\theta$, parameters of $p_G$, to minimize $\KL(p_{B|m}\,||\,p_{G|m})$, with always the same minimibatch $m$, will follow similar exploding dynamics.
\end{coro}

\paragraph{Variance over minibatches.} Note that the followed gradient $-\nabla_{p_G} \KL(p_B||p_G) = \frac{p_B}{p_G}$ is always positive, at all locations $x$, and thus at each time step, the density increases at sampled points
of the current minibatch,
and is implicitely reduced at other locations by the normalizing flow architecture. This positive pressure will cancel out when integrated over the whole space, as one cannot increase densities at all points simultaneously while keeping the total mass constant; in the end, pressures matter only relatively to the average one (densities at locations $x$ with relatively weak pressure will decrease).
In practice this induces a lot of variance, as one needs to wait for mini-batches to have covered the whole space for the average gradient to be informative, furthermore hoping that the resulting sampling will be sufficiently uniform. This slows down and may significantly hinder the optimization of quantities such as $\KL$ strongly relying on global quantities (unit mass).

\paragraph{Correct normalization.} The problem above disappears with correct normalization over mini-batches, ensuring that the distributions inside $\KL$ have unit mass:
$$p^{d|m}_{B}(x) := \frac{1}{\DDS\sum_{i\in m} p_B(x_i)} \sum_{i\in m} \delta_{x = x_i} p_B(x_i) \;\;\;\;\;\;\text{  and  }\;\;\;\;\;\; p^{d|m}_{G}(x) := \frac{1}{\DDS\sum_{i\in m} p_G(x_i)}  \sum_{i\in m} \delta_{x = x_i} p_G(x_i)$$
As $p^{d|m}_{B}$ and $p^{d|m}_{G}$ are probability distributions over minibatch points, the divergence $\KL(p_B^{d|m}\|p_G^{d|m})$ makes sense, as well as its gradient w.r.t. the parameters $\theta$ of the generative model $p_G$, as redistributing the mass within the minibatch, without pulling extra mass from non-sampled points.
Therefore each minibatch gradient is informative, and convergence is much faster.
On the opposite, the divergence $\KL(p_B^{d|m}\|p_G)$
and the former discretization of the divergence $\KL(p_B||p_G)$ over the minibatch both suffer from manipulating densities $p_G(x_i)$ that are not constrained to sum to 1 over the minibatch, leading to the previously detailed issues.

\subsection{Reducing the variance of estimators induced by discretization}

\paragraph{Losses vs.~estimators of them by discretization.}
One important thing is not to confuse a quantity, such as $Q = \KL(p_B||p_G)$, with estimators $\widehat{Q}$ of it, such as the approximations obtained by discretization over mini-batches of samples. Another important thing is not to confuse the gradient $\nabla \widehat{Q}$ of a good estimator of a quantity with a good estimator $\widehat{\nabla Q}$ of the gradient of that quantity. This is the latter one that we aim at finding, and that we study now.

We would like to estimate the gradient of $\KL(p_B\|p_G)$ (or of another similar criterion) w.r.t. the generator parameters $\theta$.

Such gradient is of the form:
$$A := \int_X g\; d\mu= \int_{x\in X} g(x)\; d\mu(x)$$
where for instance in the case of $\nabla_\theta \KL(p_B\|p_G) = \int_X \frac{p_B}{p_G} \frac{dp_G}{d\theta}$, one could choose $g(x) = \frac{ p_B(x) }{ p_G(x) } \frac{dp_G(x)}{d\theta}$ and $d\mu = dx$, or $g = \frac{ p_B }{ p^2_G } \frac{dp_G}{d\theta}$ and $d\mu(x) = p_G(x)dx$, depending on the sampler. Yet, all we can do is to sample a minibatch $m$ containing $n$ samples, chosen i.i.d. according to $d\mu$:
$$\hA := \frac{1}{n}\sum_{i \in m} g(x_i)$$
This estimator is unbiased: $\E_m[\hA] = A$, i.e. on average over all possible minibatches, $\hA$ becomes $A$.
The approximation error, or \emph{estimator variance}, can be shown to be:
$$\E_m [ (A-\hA)^2] =  \frac{1}{n} \left( \E_x \left[ g^2 \right] -  A^2 \right)$$

\paragraph{Stabilizing trick to reduce estimator variance.}
Note that adding $+ \int_X K \frac{dp_G}{d\theta}$ to the gradient $A$, for some constant $K$, does not change it, as $ \int_X \frac{dp_G}{d\theta} =  \frac{d}{d\theta}  \int_X p_G =  \frac{d}{d\theta}(1) = 0$.
%It amounts to adding $+K \int_X p_G = K$ to the optimization criterion.
A new expression of our target quantity $A$, with its associated unbiased estimator $\hA'$, is thus:
$$A = \int_X g = \int_X g + K \frac{dp_G}{d\theta}
\;\;\;\;\;\;\;
\text{and}
\;\;\;\;\;\;\;
\hA' := \frac{1}{n}\sum_{i \in m} g(x_i) + K \frac{dp_G(x_i)}{d\theta} $$
Minimizing the variance of the estimator $\hA'$  w.r.t.~$K$ yields
$K^* = -\E\left[g  \frac{dp_G}{d\theta} \right] / \E\left[ \frac{dp_G}{d\theta}^2 \right] \simeq - \sum_j g(x_j) \frac{dp_G(x_j)}{d\theta} / \sum_j(\frac{dp_G(x_j)}{d\theta})^2$
where the approximation is performed
through running means over past minibatches.

This variance-reduced estimator of the gradient can easily be obtained by just
adding $+K^* \int_X p_G$ to the optimization criterion before discretization over the minibatch.
The  corresponding gradient descent will be more robust to minibatch discretization.

\paragraph{Normalization mistakes are removed by the stabilizing trick (on average).}
Applying the trick above removes the normalization mistakes of the naive discretization. Indeed, on average, i.e. on expectation over minibatches,
one can show that the total mass within a minibatch is preserved by a gradient step based on the trick-corrected gradient estimator.
This was not the case with the naively-discretized $\KL$ gradient, which always asks for increasing the mass at each point of each minibatch.

\section{Proofs and details of the previous section}
\label{sec:appendix:pitfalls_details}

\subsection{Reminder about KL divergence}

The quantity:
$$\KL(p||q) = \int_X p \log \frac{p}{q}$$
is called a ``divergence" and has the following properties, when applied to 2 probability distributions $p$ and $q$ defined over a space $X$:
\begin{itemize}
\item $\KL(p||q) > 0$ for any different $p$ and $q$
\item $\KL(p||q) = 0$ if and only if $p = q$
\end{itemize}

This is known as Gibbs inequality and makes $\KL$ usable as a criterion to measure how far two probability distributions are from each other. $\KL$ is not a distance in the mathematical sense though. In particular,  $\KL(p||q)$ is not equal to $\KL(q||p)$ in general.

Note 1: it is important that $p$ and $q$ be \emph{probability} distributions, i.e. $p \geqslant 0$ and $\int_X p = 1$, and similarly for $q$. Without these constraints, Gibbs inequality is not true anymore, and thus minimizing $\KL(p||q)$ w.r.t. $q$ might lead to a solution $q^*$ different from $p$.

Note 2: formulas containing the symbol $\int_X$ are generically true for any measure over $X$, and not necessarily just the Lebesgue measure. For instance, one can replace $\int_X$ with $\sum_i \delta_{x=x_i}$, i.e. consider a discrete set of points $\{x_i\}$, or a weighted set: $\sum_i w_i \delta_{x=x_i}$ with $w_i \geqslant 0$.

\subsection{First pitfall: dynamics of minimizing \texorpdfstring{$\KL(p||q)$}{KL(p||q)} w.r.t. \texorpdfstring{$q$}{q} without constraining \texorpdfstring{$\int_X q$}{int q} to be 1}

Let us start from $q_{t=0} = q_0$ and minimize $\KL(p||q_t)$ by gradient descent w.r.t. $q_t$ directly (no intermediate parameterization), i.e. $\frac{dq}{dt} = - \nabla_q \KL(p||q_t)$

To obtain the expression of the functional gradient $\nabla_q \KL$, let us consider any infinitesimal variation $\delta q$ of $q$ : the quantity $\KL(p||q) = \int_X p(x) \log \frac{p(x)}{q(x)} dx = -\int_X p(x) \log q(x) dx + \mathrm{Constant}$ would change by:
$$\delta(\KL(p||q))(\delta q) = - \int_X \frac{p(x)}{q(x)} \delta q(x) dx$$
As the ($L^2$) gradient $\nabla_q f(q)$ of a function $f$ is defined as the unique distribution $v$ such that $\int_X v(x) \,\delta q(x) dx = \delta(f(q))(\delta q) + o(\delta q) \;\;\;\forall \delta q$, one has:
$$\nabla_q \KL(p||q) = -\frac{p}{q}$$ 

Hence the dynamics rewrite:
$$\frac{dq}{dt} = \frac{p}{q}$$
in the sense that $\forall x, \;\frac{dq_t(x)}{dt} = \frac{p(x)}{q_t(x)}$. 
Let us note that this implies $\frac{dq}{dt} \geqslant 0 \; \forall x, t$ and that as a consequence, $q(x)$ increases with time for all $x$, getting farther and farther away from the missing constraint $\int_X q = 1$. 
Indeed:
$$\frac{dq^2}{dt} = 2p$$
and hence:
$$q^2(t) = 2pt + q^2_0$$
$$q(t) = \sqrt{2pt + q^2_0}$$
Thus, for large times $t$,
$$q(t) \simeq \sqrt{2t} \sqrt{p}$$
in the sense that:
$$\forall x, \;\; q_t(x) \simeq \sqrt{2t} \sqrt{p(x)}$$

which shows that:
\begin{itemize}
\item  there is a lack of normalization ($q$ will get infinite mass),
\item even correcting by the total mass of $q$ will not yield $p$, but $\sqrt{p}$,
\item $q = p$ is not the minimizer of $\KL(p||q)$.
\end{itemize}

Indeed with $q_t = \sqrt{2t} \sqrt{p}$ one gets:
$$\KL(p||\sqrt{2tp}) = \frac{1}{2}H(p) - \frac{1}{2} \log (2t) \;\; <\!\!\!< \;0 = \KL(p||p)$$
for $t > \frac{1}{2}e^{H(p)}$.

%\subsubsection{Why should I care? My $p_G$ is guaranteed to sum to 1.}
\subsection{Extension to normalizing flows keeping full mass constant by design}

The above also applies to distributions discretized on a minibatch $m$: 
$$p(x) = p_{B|m}(x) := \sum_i \delta_{x = x_i} p_B(x_i)$$ 
and 
$$q(x) = p_{G|m}(x) := \sum_i \delta_{x = x_i} p_G(x_i)$$
The total mass of these distributions is not 1. It can be arbitrarily high, as $p_G(x_i)$ is a density (and not a probability: it is not constrained to be less than 1).
As a consequence, a gradient descent w.r.t. $\theta$, parameters of $p_G$, to minimize $\KL(p_{B|m}||q_{G|m})$, with always the same minimibatch $m$, will follow similar exploding dynamics.

%\subsubsection{Objection: but if I write $q = m q'$ with $\int q' = 1$...}

%A particularity of (the interesting part of) $KL$ is that it rewrites this way, for any scaling factors $\alpha, \beta \in \mathbb{R}$:
%$$KL(\alpha p||\beta q) = \alpha KL(p||q) + \alpha \log \frac{\alpha}{\beta}$$
%even though it's not meant to be defined for non-unit-mass distributions. 
%One could then think, for non-unit-mass distributions $q$, of decompising them as $q = m q'$ with the mass $m = \int_X q$ and the normalized distribution $q' = \frac{1}{m} q$. Then:
%$$KL(p||q) = KL(p||q') - \log m$$
%might make think that a gradient descent w.r.t. $q$ would send $q'$ to $p$, even though the mass explodes, which could be dealt with by renormalizing $q$ from time to time (or just at the end). We have seen that it is not the case. The reason is that a gradient descent over the quantity above w.r.t. $q$ is not the same as a gradient descent over the same quantity w.r.t. $m$ and $q'$ explicitely. The metric in the tangent space is indeed not the same one, and consequently the gradient is not, as well.

\subsection{Gradient of the estimator  vs. estimator of the gradient}
The gradient $\nabla^{L^2}_q$ with respect to the distribution $q$ does not take into the fact that  $q$ should remain a probability distribution, i.e. sum up to 1. One should \emph{project} this gradient onto the set of possible variations of $q$. This can be done for instance by considering $\nabla^{L^2}_q \KL - \int_X \nabla^{L^2}_q \KL$, i.e. removing its mean. This would definitely change the dynamics studied in the previous section.

In our implementation with ``normalizing flows'', the gradient $\nabla^{L^2}_\theta$ with respect to the parameters $\theta$ of the probability distribution $q_\theta$ does implicitely take into account the fact that $q$ should remain a probability distribution, in that by construction all $q_\theta$ are probability distributions: with a ``normalizing flow'', there is no way to escape the manifold of distributions which sum up to 1.

Note the difference between:
\begin{itemize}
\item the gradient $\nabla^{L^2}_\theta$ of $\KL(p^d\|q^d)$ between discretized distributions (which is the correct way according to the previous sections): $\sum_i \frac{d \log q(x_i)}{d\theta} \, \big( q^d(x_i) - p^d(x_i) \big)$, the first term coming from the derivative of the log of the normalizing factor $\frac{1}{\sum_i q(x_i)}$
\item and the discretization of the gradient $\nabla^{L^2}_\theta$  of $\KL(p^d\|q)$ (which is an incorrect way): $- \sum_i p^d(x_i) \frac{d \log q(x_i)}{d\theta}$.
\end{itemize}

The incorrect way misses a term, which acts as if it had a supplementary term $- \frac{1}{\sum_i q(x_i)} \sum_i \frac{dq(x_i)}{d\theta}$
which leads to be a positive additive term in a gradient \emph{descent}: all $q(x_i)$ are increased.

This is similar to the actor-critic approach in reinforcement learning, where the comparison $\big( q^d(x_i) - p^d(x_i) \big)$ improves the dynamics of the training, pushing the probability flow in the right direction (the sign indicates whether more probability mass is needed or the opposite), while without the critic $p^d$ the training would be far less stable and require much more time.

\subsection{Stabilizing trick}

First note that for any parameterized probability distribution $q = p_G = p^{(\theta)}_G$:
$$\forall \theta, \;\;\;\;\;\int_X p_G = 1$$
and consequently:
$$\frac{d}{d\theta} \int_X p_G = \int_X \frac{dp_G}{d\theta} = (0,0,0\dots 0)$$
with as many 0 as parameters $\theta$.
Thus any gradient formula of the form $\int_X \frac{dq}{d\theta} f = \int_{x \in X} \frac{dq}{d\theta}(x) f(x) dx$ satisfies:
$$\int_X \frac{dq}{d\theta} f = \int_X \frac{dq}{d\theta} (f + K) \;\;\;\;\;\forall K\in\mathbb{R}^{|\Theta|}$$ 
for any additive constant $K$ (which is a vector). Thus we can form many different estimators of $\int_X \frac{dq}{d\theta} f$ by picking a value for $K$ and discretizing $\int_X \frac{dq}{d\theta} (f + K)$ over a minibatch. What we will do in next section is to note that our gradient writes in that form ($\int_X \frac{dq}{d\theta} f$), and choose within this family of estimators the one with the least variance.

Similarly, 
$$\int_X \frac{d\log q}{d\theta} f = \int_X \frac{d\log q}{d\theta} (f + Kq) \;\;\;\forall K\in\mathbb{R}^{|\Theta|}$$ 

Note: for readability purposes, we used the abusive notation  $f + K$, which stands for $f .* (1,1,1\dots) + K$, and the product with $\frac{dq}{d\theta}$ is done coefficient-wise for $K$. To be more precise:
\begin{itemize}
\item $f(x) \in \R$, so $\frac{dq}{d\theta} f$ reads $f(x) \frac{dq(x)}{d\theta}$ : real $\times$ vector multiplication,
\item $K \in \mathbb{R}^{|\Theta|}$ is a vector (as many coefficients as parameters $\theta$) and $\frac{dq}{d\theta} K$ reads $\frac{dq}{d\theta} \;.\!*\; K = \left( \frac{dq(x)}{d\theta_j} K_j  \right)_{j}$ which is a coefficient-wise multiplication, yielding a vector of same size.
\end{itemize}

\paragraph{Implementing the trick very simply as an addition to the loss}

Instead of manipulating the gradient, this trick can be implemented directly by changing the loss to be optimised.
Indeed adding $+ \int_X \frac{dq}{d\theta} K$ to the gradient amounts to adding $+K\int_X q$ to the optimization criterion (with \emph{detached} $K$ and \emph{without} replacing $\int_X q$ by its expected value, 1).

\subsection{Minimizing the variance of the estimator of the gradient}

\subsubsection{Set-up}
Consider the case where one wants to minimize $\KL(p\|q)$ (or another criterion) w.r.t. $q$.
At some point during gradient computations we would like to compute a quantity of the form:
$$A := \int_X g = \int_{x\in X} g(x) dx$$
but all we can do is sample a minibatch $m$ containing $n$ samples, chosen i.i.d. and uniformly over $X$:
$$B := \frac{1}{n}\sum_{i \in m} g(x_i)$$

NB: this i.i.d.~and uniform hypotheses will be important in the sequel. One could imagine that points are sampled on purpose far away from each other, or from another distribution. In which case, the proves below have to be adapted.

We know that on average over all possible minibatches, $B$ becomes $A$: %$\newcommand\E{\mathbb{E}}$
$$\E_m[B] = A$$
and this can be shown by: $\E_m[B] = \frac{1}{n} \E_m\left[  \sum_{i\in m} g  \right] = \E_x\left[  g(x)  \right] = A$ as minibatch points are i.i.d.~and uniformely sampled over $X$ (see next section for details).

Yet for any minibatch, $B$ is rarely exactly $A$. In particular if $X$ is large, $n$ is small, or $g$ varies quickly, $B$ is not likely to be exactly $A$.
We thus want to study the approximation error:
$$\E_m [ (A-B)^2]$$
in order to minimize it w.r.t. the parameter $K$ above.

NB: computations below are done with integrals and samplers uniform over $X$.

\subsubsection{Minimizing the variance}
$$\E_m [ (A-B)^2] = A^2 + \E_m\left[ B^2 \right] - 2A \E_m\left[ B \right] = -A^2 + \E_m\left[ B^2 \right]$$
as $A$ does not depend on $m$.
The $-A^2$ term is constant (depends neither on $m$, nor $K$). Let us study the other term, in order to optimize it w.r.t. $K$ later:
$$B^2 = \left( \frac{1}{n}\sum_{i \in m} g(x_i) \right)^2 = \frac{1}{n^2}\left( \sum_i g(x_i)^2 + \sum_{i\neq j} g(x_i)g(x_j)\right)$$
On average over minibatches, this yields:
$$\E_m\left[ B^2 \right] = \frac{1}{n^2} \E_m\left[ \sum_i g(x_i)^2  \right] + \frac{1}{n^2} \E_m\left[ \sum_{i\neq j} g(x_i)g(x_j) \right]$$

\paragraph{Useful properties of averages over minibatches.}
As minibatches are formed with samples randomly chosen, i.i.d.~(that is, each sample is independently sampled from the other ones in the minibatch), $\E_m$ is actually $\E_{x_1 \sim \mathcal{S}} \E_{x_2 \sim \mathcal{S}} \dots \E_{x_m \sim \mathcal{S}}$ where $\mathcal{S}$ is the sampling distribution of one point, and consequently formulas such as $\E_m\left[ \sum_i f(x_i) \right]$ can be simplified as follows:
$$\E_m\left[ \sum_i f(x_i) \right] = \sum_i \E_{m} \left[ f(x_i) \right] = \sum_i \E_{x_i \sim \mathcal{S}} \left[ f(x_i) \right] = n\,\E_{x \sim \mathcal{S}}\left[ f(x) \right]$$

Similarily, formulas involving 2 variables symmetrically such as $\E_m\left[ \sum_{i\neq j} f(x_i)f(x_j) \right]$ boil down as follows:
$$\E_m\left[ \sum_{i\neq j} f(x_i)f(x_j) \right] = \sum_{i\neq j} \E_m\left[  f(x_i)f(x_j) \right] = n(n-1) \E_m\left[ f(x)f(x')\right]$$
$$= n(n-1) \,\E_{x \sim \mathcal{S}}\left[ f(x)\right]\; \E_{x' \sim \mathcal{S}}\left[ f(x')\right] = n(n-1) \,\left( \E_{x \sim \mathcal{S}}\left[ f(x)\right] \right)^2$$

\paragraph{Back to our variance minimization.}

Our quantity of interest above thus becomes:
$\E_m\left[ B^2 \right] =  \frac{1}{n}\E_{x \sim \mathcal{S}}\left[ g(x)^2 \right] + \frac{n(n-1)}{n^2}\left( \E_{x \sim \mathcal{S}}\left[ g(x)\right] \right)^2$
$= \frac{1}{n}\E\left[ g(x)^2 \right] + (1-\frac{1}{n}) A^2$ in our case where the sampling distribution is uniform. % **(check what happens for other sampling distributions)**

Thus the approximation error is:
$$\E_m [ (A-B)^2] = -\frac{1}{n} A^2 + \E_m \left[ \frac{1}{n^2}\sum_i g(x_i)^2 \right]$$
which we can estimate by sampling as:
$$ \E_m [ (A-B)^2] \approx -\frac{1}{n} A^2 + \frac{1}{n^2}\sum_i g(x_i)^2$$
using as many samples as possible (not the current minibatch considered at that gradient descent step, but a sliding average over past minibatches for instance). Using the current minibatch might lead to a biased estimator.

Let us develop the second term (the first one being constant).
The target quantity $A$ is the gradient of the optimization criterion, of the form 
$$A = \int_X g = \int_X \frac{dq}{d\theta}(f + K)$$
thus 
$$g(x_i) = \frac{dq(x_i)}{d\theta} (f(x_i) + K)$$
Note that $g$ and $K$ are vectors, but we can deal with each coordinate independently as they do not interact in these expressions. Let us focus on the $j$-th coordinate of $g$, i.e. the $j$-th parameter:
$$g_j(x_i) = \frac{dq(x_i)}{d\theta_j} (f(x_i) + K_j)$$
In order not to hamper the reading, we drop $j$ in the next lines:
$$\sum_i g(x_i)^2 = \sum_i \frac{dq(x_i)}{d\theta}^2 (f(x_i)^2 + K^2 + 2Kf(x_i))$$
$$= K^2 \left(\sum_i \frac{dq(x_i)}{d\theta}^2\right) + 2K \left(\sum_i \frac{dq(x_i)}{d\theta}^2 f(x_i)\right) + \left(\sum_i \frac{dq(x_i)}{d\theta}^2 f(x_i)^2\right)$$

Minimizing this w.r.t. $K$ yields:
$$K = -\frac{\sum_i (\frac{dq(x_i)}{d\theta})^2 f(x_i)}{\sum_i(\frac{dq(x_i)}{d\theta})^2}$$
i.e.
$$K_j = -\frac{\sum_i (\frac{dq(x_i)}{d\theta_j})^2 f(x_i)}{\sum_i(\frac{dq(x_i)}{d\theta_j})^2}$$

This quantity can be efficiently computed in practice using libraries such as BackPACK\footnote{\url{https://backpack.pt/}}~\cite{dangel2020backpack}.

\subsubsection{Gradient estimation}

Thus, given a minibatch, the estimation of a gradient $A$ of the form $\int_X \frac{dq}{d\theta} f$, 
such as $\nabla_\theta \KL(p||q_\theta) = -\int_X \frac{dq}{d\theta} \frac{p}{q}$, 
should rather be done with the formula:
$$A_j = \frac{1}{n} \sum_i \frac{dq(x_i)}{d\theta_j} \left( f(x_i) -\frac{\sum_k (\frac{dq(x_k)}{d\theta_j})^2 f(x_k)}{\sum_k(\frac{dq(x_k)}{d\theta_j})^2} \right) \in \mathbb{R}$$
$$= \frac{1}{n} \left( \sum_i \frac{dq(x_i)}{d\theta_j} f(x_i) - \left( \sum_i \frac{dq(x_i)}{d\theta_j} \right) \frac{\sum_k (\frac{dq(x_k)}{d\theta_j})^2 f(x_k)}{\sum_k(\frac{dq(x_k)}{d\theta_j})^2} \right)$$
The full gradient $A$ with all coordinates is then estimated as:
$$\hat{A} = \frac{1}{n} \left( \sum_i \frac{dq}{d\theta} f - \left( \sum_i \frac{dq}{d\theta} \right) \frac{\sum_k (\frac{dq}{d\theta})^2 f}{\sum_k(\frac{dq}{d\theta})^2} \right)$$
using coefficient-wise squaring, multiplication and division between parameter-size vectors.\\ 

NB: as said above, the terms coming from $K$, i.e.~the sums involving squares, should be computed with running means over minibatches, while the other sums are performed on the current minibatch.

Remember that $\sum_i \frac{dq}{d\theta}$ is a quantity which on average over possible minibatches is 0, but is not necessarily exactly 0 for any given minibatch. One could correct the deviation of $\sum_i \frac{dq}{d\theta}$ from 0 by removing the mean of $\frac{dq}{d\theta}$, which would lead to an expression of the form:
$$\frac{1}{n} \sum_i \left(\frac{dq}{d\theta} - \overline{\frac{dq}{d\theta}}\right)  f = 
\frac{1}{n} \sum_i \frac{dq}{d\theta} f - \frac{1}{n^2}  \left(\sum_i \frac{dq}{d\theta}\right)  \left(\sum_i f \right)$$
but it turns out that with our stabilizing trick, a better correction can be found, by exploiting the correlation between $f$ and $(\frac{dq}{d\theta})^2$.

Note also that it is not useful to use this trick on normalized discretized distribution optimization such as $\KL(p^d||q^d)$. Indeed, $\sum_i \frac{dq^d}{d\theta}$ would be 0 always, so no correction would be brought.

\subsubsection{Variance}

The expected deviation between the true gradient $A$ and the (optimized) minibatch estimation $B$ is then (estimated over a minibatch):
$$\E_m [ (A-B)^2] = -\frac{1}{n} A^2 + \frac{1}{n^2}\left[ \left( \sum_i \frac{dq}{d\theta}^2 f^2 \right) - \frac{\left( \sum_i (\frac{dq}{d\theta})^2 f \right)^2}{\sum_i(\frac{dq}{d\theta})^2} \right]$$
Note that the term between brackets is positive (or 0), according to Cauchy-Schwartz. It is 0 when (and only when) $f$ is constant (in which case $A$ is 0 also).\\ %, in case you wondered about negative variances...).\\
Note also that without optimization upon $K$, i.e. with $K=0$, the negative term in the bracket disappears. Consequently, the variance reduction due to this stabilizing trick is $\frac{1}{n^2} \frac{\left( \sum_i (\frac{dq}{d\theta})^2 f \right)^2}{\sum_i(\frac{dq}{d\theta})^2}$. 

\subsection{Normalization mistakes are removed by the stabilizing trick (on average)}
We show here that applying the stabilizing trick correctly (i.e. with running means to estimate $K$) does remove the normalization mistakes of the naive discretization. Indeed, on average, i.e. on expectation over minibatches, the total mass is preserved at sampled points, as follows.

Considering the mass (density) $q(x_i)$ at point $x_i$,
the mass change during this time step, at point $x_i$, is $\frac{dq}{d\theta}(x_i) \cdot \varepsilon \,\delta\theta$ where $ \varepsilon$ is the learning rate and  $\delta\theta$ is
the parameter change, given by $\delta\theta = \sum_i \frac{dq}{d\theta} f + \sum_i \frac{dq}{d\theta} K$.

The global mass change over all sampled points is thus: 
$$\varepsilon \left( \sum_i \frac{dq}{d\theta}(x_i) \right) \cdot \left(\sum_i \frac{dq}{d\theta} f + \sum_i \frac{dq}{d\theta} K  \right) \;\;\;\;\propto \;\;\;\; \left( \sum_i \frac{dq}{d\theta} \right)\left(\sum_i \frac{dq}{d\theta} f \right) +  \left( \sum_i \frac{dq}{d\theta} \right)^2 K$$
On average over minibatches, this becomes:
  $$\E \left[ \frac{dq}{d\theta}^2 f \right] + \E \left[ \frac{dq}{d\theta}^2 \right] K$$
  which can be enlightened by the value of $K$ obtained in previous section:
  $K = -\frac{\E \left[ \frac{dq}{d\theta}^2 f\right]}{ \E \left[ \frac{dq}{d\theta}^2\right]}$.
Consequently on average the mass is kept. This prevents the pathological dynamic behavior observed with naive discretization.

\end{document}